%% file: main.tex
\documentclass{article}

\usepackage{microtype}
\usepackage{graphicx}
\usepackage{subfigure}
\usepackage{booktabs} % for professional tables
\usepackage{multirow}
\usepackage{amssymb}% http://ctan.org/pkg/amssymb
\usepackage{pifont}% http://ctan.org/pkg/pifont
\usepackage[T1]{fontenc}
\usepackage{ragged2e}

\newcommand{\xmark}{\ding{55}}%

\usepackage{hyperref}

\usepackage[accepted]{icml2025}

\usepackage{amsmath}
\usepackage{amssymb}
\usepackage{mathtools}
\usepackage{amsthm}
\usepackage{makecell}

\usepackage[capitalize,noabbrev]{cleveref}

\theoremstyle{plain}

\theoremstyle{definition}

\theoremstyle{remark}

\usepackage[textsize=tiny]{todonotes}
\usepackage{lineno}

\icmltitlerunning{Test-Time Tuned Language Models Enable End-to-end De Novo Molecular Structure Generation from MS/MS Spectra}

\begin{document}

% \twocolumn[
% ]

\onecolumn

% TITLE OPTIONS
% another option: Generating De Novo Molecular Structures from MS/MS Spectra with End-to-End Active Fine-Tuning of Language Models 
% bold option: Test-Time Fine-Tuned Language Models Enable End-to-end De Novo Molecular Structure Generation from MS/MS Spectra
% milder: Generating De Novo Molecular Structures from MS/MS Spectra with Language Models Tuned at Test-time
\icmltitle{Test-Time Tuned Language Models Enable End-to-end De Novo Molecular Structure Generation from MS/MS Spectra}

\icmlsetsymbol{equal}{*}

\begin{icmlauthorlist}
\icmlauthor{Laura Mismetti}{ibm,eth,nccr}
\icmlauthor{Marvin Alberts}{ibm,nccr,uzh}
\icmlauthor{Andreas Krause}{eth}
\icmlauthor{Mara Graziani}{ibm,nccr}
\end{icmlauthorlist}

\icmlaffiliation{ibm}{IBM Research, Säumerstrasse 4, 8803 Rüschlikon, Switzerland}
\icmlaffiliation{eth}{Department of Computer Science, ETH Zürich, 8092 Zürich, Switzerland}
\icmlaffiliation{nccr}{NCCR Catalysis, Switzerland}
\icmlaffiliation{uzh}{University of Zürich, Department of Chemistry, 11, Winterthurerstrasse 190, 8057 Zürich, Switzerland}

\icmlcorrespondingauthoremailonly{laura.mismetti1@ibm.com}
\icmlcorrespondingauthoremailonly{mara.graziani@ibm.com}

\printAffiliationsTop{}

\icmlkeywords{Machine Learning, Transformers, Tandem Mass Spectroscopy, MS/MS, Structure Elucidation}

\vskip 0.6in

% \linenumbers

\begin{abstract}
% I think it is better to rephrase some parts to highlight more the method, which could be a solution to the out-of-distribution problem, rather than the performances on the general structure elucidation from MSMS task!!
% SIRIUS relies on PubChem database containing ~123M fingerprints (equal molecules) - while in our method fingerprints are not used and only a far lower amount of compounds is used for training.
\vspace{0.5cm}
Tandem Mass Spectrometry is a cornerstone technique for identifying unknown small molecules in fields such as metabolomics, natural product discovery and environmental analysis. 
However, certain aspects, such as the probabilistic fragmentation process and size of the chemical space, make structure elucidation from such spectra highly challenging, particularly when there is a shift between the deployment and training conditions.
% further constrained by differences between the training data and the target experimental conditions, which reduce the transferability of the models to real everyday use where unseen compounds are encountered frequently. 
Current methods rely on database matching of previously observed spectra of known molecules and multi-step pipelines that require intermediate fingerprint prediction or expensive fragment annotations. 
% Moreover, the task remains particularly challenging for compounds absent from reference databases. 
We introduce a novel end-to-end framework based on a transformer model that directly generates molecular structures from an input  tandem mass spectrum and its corresponding molecular formula, thereby eliminating the need for manual annotations and intermediate steps, while leveraging transfer learning from simulated data.
To further address the challenge of out-of-distribution spectra, we introduce a test-time tuning strategy that dynamically adapts the pre-trained model to novel experimental data. 
Our approach achieves a Top--1 accuracy of
% and sets the new state of the art to 
3.16\% on the MassSpecGym benchmark and 12.88\% on the NPLIB1 datasets, considerably outperforming conventional fine-tuning. Baseline approaches are also surpassed by 27\% and 67\% respectively. %on MassSpecGym dataset.
%
%Our method surpasses the de-facto state-of-the-art approach DiffMS, by 67\% on NPLIB1 and 27\% on MassSpecGym.
Even when the exact reference structure is not recovered, the generated candidates are chemically informative, exhibiting high structural plausibility as reflected by strong Tanimoto similarity to the ground truth. Notably, we observe a relative improvement in average Tanimoto similarity of 83\% on NPLIB1 and 64\% on MassSpecGym compared to state-of-the-art methods.
Our framework combines simplicity with adaptability, generating accurate molecular candidates that offer valuable guidance for expert interpretation of unseen spectra. 
\vfill
\end{abstract}

%% GRAPHICAL ABSTRACT
% \begin{figure*}[h]
\begin{minipage}[t]{\linewidth}
\centering
% \vspace{-20px}
\includegraphics[width=\textwidth]{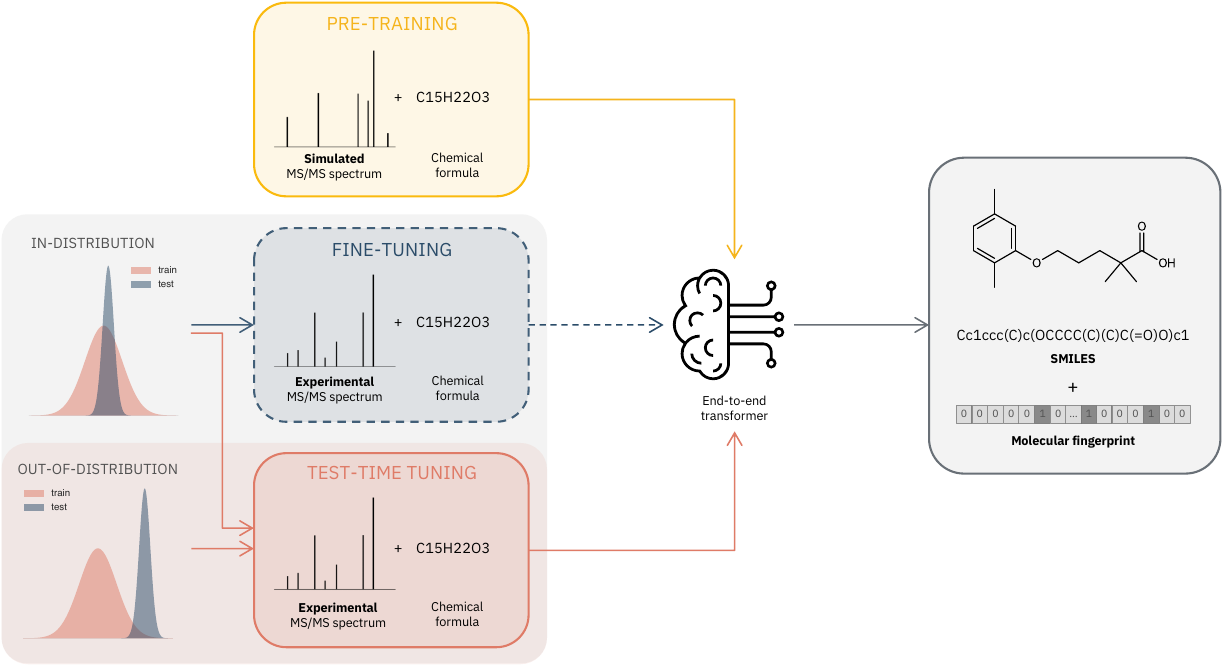} \\
\vspace{0.5cm}
% \caption{
\justifying
\noindent \textbf{Graphical abstract}\ \ We propose a transformer encoder–decoder model that predicts SMILES and molecular fingerprints from a MS/MS spectra and its associated chemical formula. The model is pre-trained on simulated spectra~\cite{alberts_unraveling_2024} and adapted via test-time tuning on experimental data from NPLIB1~\cite{duhrkop_systematic_2021} and MassSpecGym~\cite{bushuiev_massspecgym_2024}. We compare standard fine-tuning to our proposed test-time tuning strategy, which selects the training samples that are the most informative about the target experimental spectra based on the prediction of molecular fingerprints, in in-distribution and out-of-distribution scenarios.
% }
% \label{fig:initial_schema}
% \end{figure*}
\end{minipage}
\onecolumn
\justifying

\newpage
\section{Introduction}\label{sec:intro}
\input{sections/introduction}

% NOTE: moving methods after introd. as all journals have structure Intro, Results, Discussion, Methods, References

\section{Test-time tuning for adaptation to experimental conditions}
\input{sections/test-time-tuning}

\section{Results}\label{sec:results}
\input{sections/results}

% \subsection{Environmental application}\label{sec:env-app}
% \input{sections/environmental-application}

% \newpage
\section{Discussion}\label{sec:discussion}
\input{sections/discussion}

\section{Methods and data}\label{sec:methods-data}
\input{sections/methods}

\section*{Data Availability}
All the data used in this study are public. In particular, simulations dataset can be downloaded from \url{https://rxn4chemistry.github.io/multimodal-spectroscopic-dataset/}. The experimental dataset MassSpecGym can be found at \url{https://huggingface.co/datasets/roman-bushuiev/MassSpecGym}, while NPLIB1 can be retrieved at \url{https://bio.informatik.uni-jena.de/wp-content/uploads/2020/08/svm_training_data.zip} and pre-processed following the procedure provided in Appendix \ref{sec:appendix-data}.

\section*{Code Availability}
The code used in this study is open-source (MIT license) and can be found as a GitHub repository at \url{https://github.com/rxn4chemistry/MultimodalAnalytical/tree/ttt-msms}.

\newpage
\bibliography{biblio_msms}
\bibliographystyle{icml2025}

\section*{Acknowledgements}
This publication was created as part of NCCR Catalysis (grant number 225147), a National Centre of Competence in Research funded by the Swiss National Science Foundation.

\section*{Author Contributions}
L.M., M.A., M.G. conceived the project. L.M. wrote the software, conducted the experiments and wrote the manuscript. M.A., M.G., A.K. supervised the work. 

\section*{Competing interests}
The authors declare no competing interests.

\section*{Impact statement}
This work introduces a simple end‑to‑end transformer pipeline for de novo structure elucidation from MS/MS spectra of small molecules and a test‑time tuning strategy for out‑of‑distribution adaptation. The approach sets state‑of‑the‑art results on NPLIB1 (12.88\% Top‑1) and MassSpecGym (3.16\% Top‑1), surpassing DiffMS on filtered benchmarks.
Even when exact matches are not obtained, predictions remain chemically meaningful with high Tanimoto similarity and low MCES, offering interpretable candidates that strengthen expert‑guided identification of unknown compounds.
To our knowledge, this is the first application of test‑time tuning to de novo small‑molecule structure generation from MS/MS spectra, extending TTT beyond prior proteomics use cases.

\newpage
\include{sections/appendix}

\end{document}

%% file: sections/introduction.tex
Deciphering the molecular structure of an unknown compound from spectroscopic data is one of the most challenging puzzles in analytical chemistry, where Nuclear Magnetic Resonance (NMR), Infrared (IR) and Tandem Mass (MS/MS) spectroscopy~\cite{weatherly_heuristic_2005, kapp_overview_2007} provide complementary evidence to reconstruct the underlying structure. 
Classical workflows accelerate discovery via heuristic database matching against known references~\cite{duhrkop_sirius_2019,wang_masst_2020,li_toplib_2025,csi_fingerid}, yet they struggle to scale with the combinatorial breadth
of chemical diversity and the effectively unbounded chemical space.
%However, the manual annotation and analysis of spectroscopic data often is cumbersome and slow process, representing a bottleneck in the structure elucidation workflow.
% To accelerate this process, several heuristic and database search based methods have been developed to match experimentally observed spectra to  known references~\cite{duhrkop_sirius_2019, wang_masst_2020,li_toplib_2025, csi_fingerid}, yet they struggle to scale with the combinatorial breath of chemical diversity and the effectively unbounded chemical space.
%Even though widely used these methods face a major limitation: they cannot scale across the infinite chemical space and thus fail to generalise to compounds which are not present in the reference database. 
% This challenge is pronounced for MS/MS, where variations in instrumentation and acquisition parameters introduce high spectral variability in the observed spectra, further complicating database matching. 
Artificial Intelligence (AI) is increasingly effective for de-novo generation in Chemistry~\cite{schwaller2019molecular,born2023regression,frieder2023mathematical}, and has demonstrated potential for structure elucidation from NMR~\cite{jonas_deep_2019, sridharan_deep_2022, alberts_learning_2023, schilter_unveiling_2023, hu_accurate_2024, devata_deepspinn_2024, alberts_phosphor_2025}, IR~\cite{fine_spectral_2020, enders_functional_2021, alberts_leveraging_2024, alberts_setting_2025, wu_transformer-based_2025}, and multimodal approaches combining diverse spectra simultaneously~\cite{priessner_enhancing_2024, alberts_automated_2025}.

Existing methods to perform structure elucidation from MS/MS spectra either rely on expert-curated fragment annotations or learn spectral fingerprints, both of which limit the applicability to novel compounds ~\cite{spec2vec, huber_ms2deepscore_2021}. 
Other approaches predict molecular structures either passing through molecular fingerprints~\cite{goldman_mist-cf_2024, csi_fingerid} or directly from spectra~\cite{spec2mol, butler_ms2mol_2023,shrivastava_massgenie_2021,wang_madgen_2025}.
% ~\cite{wolf_silico_2010, ridder_automatic_2014, ruttkies_metfrag_2016,csi_fingerid, goldman_mist-cf_2024,spec2mol, butler_ms2mol_2023, shrivastava_massgenie_2021, wang_madgen_2025,stravs_msnovelist_2022,bohde_diffms_2025,spec2vec, huber_ms2deepscore_2021}.
However, robust generalization remains a bottleneck.
The MassSpecGym benchmark~\cite{bohde_diffms_2025} makes this complexity explicit: despite substantial methodological diversity, reported Top--1 accuracies remain low, underscoring the persistent challenge of reliable structure reconstruction from MS/MS alone. 
% ~\cite{goldman_mist-cf_2024,csi_fingerid,spec2mol, butler_ms2mol_2023, shrivastava_massgenie_2021, wang_madgen_2025}. 
%} attempt to predict molecular structures either passing through molecular fingerprints~\cite{
%} or directly from spectra~\cite{spec2mol, butler_ms2mol_2023, shrivastava_massgenie_2021, wang_madgen_2025}.
Among leading approaches, MSNovelist casts the problem as a sequence-to-sequence task that generates SMILES~\cite{stravs_msnovelist_2022}, 
MADGEN first retrieves the molecular scaffold, which is then used as starting point for a generative model~\cite{wang_madgen_2025},
whereas DiffMS first learns molecular fingerprints and then performs iterative diffusion-based reconstruction, yielding the strongest reported MassSpecGym Top--1 accuracy to date at 2.30\%~\cite{bohde_diffms_2025}. 
The main challenge of the MassSpecGym dataset lies in the domain shift between the training and test sets. Molecules present in the test set exhibit considerable differences from those found in the training set,
an important challenge that standard fine-tuning leaves unaddressed (c.f. Figure~\ref{fig:analysis-datasets}). 
This discrepancy matches realistic expectations of structure elucidation models, where the spectra used in real-world applications can differ substantially from the reference data used for training. 

\begin{figure*}[ht]
    \centering
    \vspace{0.5cm}
    \includegraphics[width=0.95\textwidth]{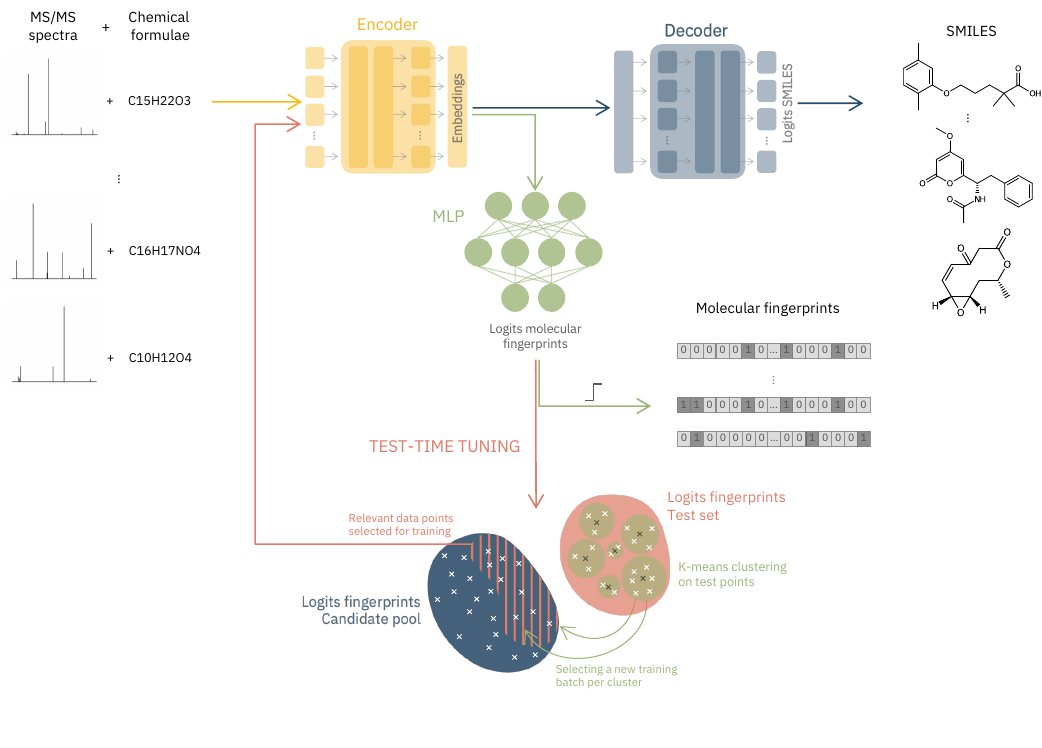}
    \vspace{-0.7cm}
    \caption{Schematic illustration of test-time tuning workflow: MS/MS spectrum and chemical formula are the input of the transformer encoder–decoder which predicts SMILES. The encoder generates embeddings used as input to a multilayer perceptron (MLP) trained to predict molecular fingerprints through an additional loss term. The logits produced by the MLP are the projection into a chemical feature space, and this representation is used to identify and select the most relevant training samples from the candidate pool for adaptation. In particular, the fingerprints of the test points are predicted (starting only from spectrum and chemical formula as input) and K-means clustering is performed on them. The selection is performed using one point per cluster at a time, depending on the cosine similarity of the fingerprints logits. The selected samples are then used for gradient updates. This process is repeated until all the clusters have been used to select a new training batch.}
    \label{fig:schema_TTT}
\end{figure*}

Transductive learning is a technique widely used in other fields to mitigate this issue~\cite{book_transduction,farahani_brief_2021}. It leverages unlabeled target-domain samples to adapt the model at inference time by selecting the most informative points from a candidate pool—typically the available training set—and training only on these selected samples. 
% This perspective follows the test-time training paradigm introduced for robustness under distribution shifts~\cite{sun2020test} and extended with informative data selection at test-time~\cite{hubotterefficiently}, and 
We evaluate whether this common domain adaptation strategy can successfully narrow the domain gap observed in Tandem Mass spectroscopy (see Figure \ref{fig:schema_TTT}). 
Since each target experimental spectrum is available, and only its structure is unknown, it can be used as an unlabeled sample to adapt the model at test time~\cite{sun2020test} to be more robust to eventual shifts in the test distribution~\cite{hubotterefficiently}. 
Crucially, while test-time training has been applied in proteomics for MS/MS spectrum prediction~\cite{pept3}, it has not been applied to de novo small-molecule structure generation, which is the task we target here.

% Test-time tuning is made possible by our architecture, which is detailed in Figure~\ref{fig:schema_TTT} and Section~\ref{sec:methods-ttt}, and that uses a transformer encoder–decoder to generate molecular structures and at the same time predict molecular fingerprints. The latter are key to identify within the training data which samples are the most informative about the experimental target data at test time and should therefore be used for tuning the model parameters.

% To the best of our knowledge, this is the first work to explore the integration of this approach with test-time tuning~\cite{hubotterefficiently} in the context of MS/MS or spectroscopic techniques. We show that this combination can effectively guide the learning toward the identification of novel, unseen compounds, leading to substantial performance improvements. The details of this approach are presented in Figure~\ref{fig:schema_TTT} and Section~\ref{sec:methods-ttt}.

We introduce a novel framework for structure elucidation from MS/MS spectra that, unlike existing approaches~\cite{stravs_msnovelist_2022,bohde_diffms_2025}, eliminates the need for intermediate annotations or predicted fragments and reports consistent accuracy improvements across all benchmarks. 
To address variability across datasets, we explore two adaptation strategies: classical fine-tuning and test-time tuning. 
% Simulated spectra~\cite{alberts_unraveling_2024} were also leveraged to improve structural consistency and address variability across datasets. 
Our method builds on a transformer encoder–decoder architecture, pre-trained on a large corpus of simulated spectra~\cite{alberts_unraveling_2024} and leverages predicted molecular fingerprints to improve structural consistency. The latter are key in the test-time tuning strategy to identify within the training data which samples are the most informative about the experimental target data and should therefore be used for tuning the model parameters. 
Compared to traditional finetuning, test‑time tuning is particularly effective on the out‑of‑distribution spectra in MassSpecGym, and standard fine‑tuning suffices when the train and test distributions are aligned.
%This adaptive tuning can effectively guide the learning toward the identification of novel, unseen compounds, unlocking stronger cross-domain performance and achieving competitive results on the NPLIB1 dataset and the MassSpecGym benchmark~\cite{duhrkop_systematic_2021,bushuiev_massspecgym_2024}. 
Even when predictions deviate from the reference structure, our generated candidates remain chemically informative, compared to those obtained from existing methods, providing valuable guidance to make an informed guess about the compound.
These results demonstrate that our framework has the potential to substantially streamline structure elucidation routines from MS/MS spectra, facilitating its integration into high-throughput workflows where rapid and accurate identification of unknown compounds is essential. 

%% file: sections/test-time-tuning.tex
Test‑time tuning is an adaptation strategy~\citep{sun2020test,hubotterefficiently} that refines a model during prediction by leveraging unlabeled test points~\citep{book_transduction,farahani_brief_2021}.
Rather than relying solely on knowledge acquired during pre-training, the algorithm uses each test point to identify a small set of relevant examples from a larger labeled dataset—typically the training set—known as the candidate pool. The model is then updated using only these selected examples, improving its understanding of the unknown test point.
In our case, the goal is to recover information about an unknown molecule. Since MS/MS spectra from the same molecule can vary significantly, the selection process relies on approximate molecular structure information rather than spectral similarity. For each test spectrum, the model first computes a molecular fingerprint, which is then compared to the fingerprints of all spectra in the candidate pool. The most similar entries—usually identified through a nearest-neighbor search—are selected. These examples often share structural features with the unknown molecule, providing valuable information for refining the model.
By iterating this selection and update process across the test set, the model gradually adapts to the specific characteristics of the new data. A detailed description of the retrieval step, clustering procedure, and update mechanism is provided in Section \ref{sec:methods-ttt}.

%% file: sections/results.tex
\subsection{Consistent improvements over \textit{de-facto} state-of-the-art methods
% and standard fine-tuning
}\label{sec:results-outperforming-sota}
We benchmark our approach comparing test-time tuning (TTT), standard fine-tuning (FT) and pre-training (PT) performances on the NPLIB1~\cite{duhrkop_systematic_2021} and MassSpecGym~\cite{bushuiev_massspecgym_2024} datasets across multiple evaluation metrics (c.f. Table~\ref{tab:results-sota}).
The proposed framework delivers consistent improvements in Top--$k$ accuracy compared to existing models, highlighting its ability to generalize across diverse datasets and domain conditions.
In particular, state-of-the-art performances are achieved on both datasets, with a Top--1 accuracy of 12.88\% on NPLIB1 and 3.16\% on MassSpecGym, on which all the baselines presented in \cite{bushuiev_massspecgym_2024} only obtained 0\%.
For completeness, we make a distinction between the original NPLIB1 dataset proposed by~\cite{duhrkop_systematic_2021}, which we refer to as {NPLIB1-Full}, and a smaller version, here named {NPLIB1-DiffMS}, obtained by following the preprocessing procedure proposed in ~\citet{bohde_diffms_2025} (see Section \ref{sec:methods-data} for further details).
Our framework also surpasses the de-facto state-of-the-art DiffMS, with a relative gain of {67\% in Top--1 accuracy on NPLIB1-DiffMS and 27\% on MassSpecGym-DiffMS}, with Top--1 accuracy of 13.95\% and 2.93\% respectively.\\
Beyond accuracy, the Tanimoto similarity and Maximum Common Edge Subgraph (MCES) distance in Table \ref{tab:results-sota} also demonstrate that our approach consistently generates candidates that are chemically closer to the ground truth, providing richer structural insights and supporting more informed decision-making during the elucidation process than the existing counterparts. An in-depth analysis of the predictions is presented in Section \ref{sec:results-similar-predictions}.
In many analytical and predictive workflows, newly acquired data, such as newly measured MS/MS spectra, are measured without substantial prior knowledge of the underlying chemical distribution. Consequently, it is inherently unclear whether a given test instance lies within the distribution represented in the model’s training data or departs substantially from it. This uncertainty motivates the development of methods capable of performing reliably under both conditions—when test data are in‑distribution (i.e., sufficiently similar to training examples) and when they are out‑of‑distribution. In Sections \ref{sec:results-TTT-case1} and \ref{sec:results-TTT-case2} we demonstrate that the proposed test‑time tuning framework is robust across both regimes, while comparing with conventional fine‑tuning.

% \vspace{0.5cm}

% \vspace{0.2cm}
\begin{table}[ht]
\centering
\resizebox{\textwidth}{!}{%
\begin{tabular}{lcccccc}
\toprule 
\multirow[c]{2}{*}{MODEL} & \multicolumn{3}{c}{Top--1} & \multicolumn{3}{c}{Top--10} \\
\cmidrule(l{0.15cm}r{0.15cm}){2-4}
\cmidrule(l{0.15cm}r{0.15cm}){5-7}
 & ACCURACY $\uparrow$ & MCES $\downarrow$ & TANIMOTO $\uparrow$ & ACCURACY $\uparrow$ & MCES $\downarrow$ & TANIMOTO $\uparrow$ \\
% \midrule
\toprule 
\multicolumn{7}{c}{\textsc{NPLIB1-Full}} \\
\midrule 
% Spec2Mol~\cite{spec2mol} & 0.00\% & 27.82 & 0.12 & 0.00\% & 23.13 & 0.16 \\
% MADGEN ~\cite{wang_madgen_2025}& 1.0\% & 70.45 & - & 1.0\% & 45.64 & - \\
% MIST + Neuraldecipher~\cite{goldman_mist-cf_2024,le_neuraldecipher_2020}& 2.32\% & 12.11 & 0.35 & 6.11\% & 9.91 & 0.43 \\
% MIST + MSNovelist~\cite{goldman_mist-cf_2024,stravs_msnovelist_2022} & 5.40\% & 14.52 & 0.34 & 11.04\% & 10.23 & 0.44 \\
% DiffMS ~\cite{bohde_diffms_2025}& 8.34\% & 11.95 & 0.35 & 15.44\% & 9.23 & 0.47 \\
{This work (TTT)}& {12.21\%}& 6.65 & 0.59 & {28.80\%}& 4.67 & 0.74 \\
{This work (FT)} & {12.42\%}& {6.40} & {0.61} & {30.83\%}& \textbf{4.42} & \textbf{0.76} \\
\textbf{This work (TTT-FT)}& \textbf{12.88\%}& \textbf{6.28} & \textbf{0.62} & \textbf{31.32\%}& \textbf{4.42} & \textbf{0.76} \\
% \textcolor{red}{This work (Extended FT)}& {7.16\%}& & & {34.67\%}& \textbf{}& \textbf{}\\
% \textcolor{red}{This work (Extended TTT)}& \textcolor{red}{9.89\%}& \textbf{} & \textbf{} & \textcolor{red}{22.90\%}& {} & \textbf{} \\
\midrule
\multicolumn{7}{c}{\textsc{NPLIB1-DiffMS}} \\
\midrule 
Spec2Mol~\cite{spec2mol}$\ddagger$ & 0.00\% & 27.82 & 0.12 & 0.00\% & 23.13 & 0.16 \\
MADGEN ~\cite{wang_madgen_2025}$\dagger$ & 1.0\% & 70.45 & - & 1.0\% & 45.64 & - \\
MIST + Neuraldecipher~\cite{goldman_mist-cf_2024,le_neuraldecipher_2020}$\ddagger$ & 2.32\% & 12.11 & 0.35 & 6.11\% & 9.91 & 0.43 \\
MIST + MSNovelist~\cite{goldman_mist-cf_2024,stravs_msnovelist_2022}$\ddagger$ & 5.40\% & 14.52 & 0.34 & 11.04\% & 10.23 & 0.44 \\
DiffMS ~\cite{bohde_diffms_2025}$\ddagger$ & 8.34\% & 11.95 & 0.35 & 15.44\% & 9.23 & 0.47 \\
{This work (FT)} & {13.20\%}& {5.77} & \textbf{0.64} & {31.14\%}& {3.93} & \textbf{0.77} \\
\textbf{This work (TTT)}& \textbf{13.95\%}& \textbf{6.07}& 0.63& \textbf{31.38\%}& \textbf{3.96}& \textbf{0.77}\\
% \textcolor{red}{This work (TTT-FT)} &&&&&&\\
% \textcolor{red}{This work (Extended FT)}& {6.10\%}& & & {42.34\%}& \textbf{}& \textbf{}\\
% \textcolor{red}{This work (Extended TTT)}& \textcolor{red}{7.23\%}& \textbf{} & \textbf{} & \textcolor{red}{24.66\%}& & \textbf{} \\
\midrule 
\multicolumn{7}{c}{\textsc{MassSpecGym}} \\
\midrule
SMILES Transformer~\cite{smiles_transformer, smiles_transformer2}$^*$ & 0.00\% & 79.39 & 0.03 & 0.00\% & 52.13 & 0.10 \\
SELFIES Transformer~\cite{selfies_transformer}$^*$ & 0.00\% & 38.88 & 0.08 & 0.00\% & 26.87 & 0.13 \\
Random Generation~\cite{bushuiev_massspecgym_2024}$^*$ & 0.00\% & 21.11 & 0.08 & 0.00\% & 18.26 & 0.11 \\
% {This work (FT)} & {1.13\%} & {}& {}& {2.93\%} & {}& {}\\
\textbf{This work (TTT)} & \textbf{3.16\%}& \textbf{11.77}& \textbf{0.46}& \textbf{6.07\%}& \textbf{9.65}& \textbf{0.54}\\
\midrule
\multicolumn{7}{c}{\textsc{MassSpecGym-DiffMS}} \\
\midrule
MIST + MSNovelist~\cite{goldman_mist-cf_2024,stravs_msnovelist_2022}$\ddagger$ & 0.00\% & 45.55 & 0.06 & 0.00\% & 30.13 & 0.15 \\
Spec2Mol~\cite{spec2mol}$\ddagger$ & 0.00\% & 37.76 & 0.12 & 0.00\% & 29.40 & 0.16 \\
MIST + Neuraldecipher~\cite{goldman_mist-cf_2024,le_neuraldecipher_2020}$\ddagger$ & 0.00\% & 33.19 & 0.14 & 0.00\% & 31.89 & 0.16 \\
MADGEN~\cite{wang_madgen_2025}$\dagger$ & 0.8\% & 74.19 & - & 1.6\% & 53.50 & - \\
DiffMS~\cite{bohde_diffms_2025}$\ddagger$ & 2.30\% & 18.45 & 0.28 & 4.25\% & 14.73 & 0.39 \\
% {This work (FT)} & {1.09\%} & & {}& {2.87\%} & {}& {}\\
\textbf{This work (TTT)} & \textbf{2.93\%} & \textbf{11.81}& \textbf{0.46}& \textbf{5.51\%} & \textbf{9.75}& \textbf{0.53}\\
\bottomrule
\end{tabular}
}
\caption{De novo structural elucidation performance on NPLIB1 \cite{duhrkop_systematic_2021} and MassSpecGym \cite{bushuiev_massspecgym_2024} datasets, and DiffMS versions of the same (limited to compounds containing only carbon, oxygen, nitrogen, hydrogen, chlorine, fluorine, sulfur or phosphorus atoms and spectra obtained with H$^+$ adduct). 
The best performing model for each metric is highlighted in bold.\\
% \underline{Underlined} the results obtained performing database search on PubChem~\cite{pubchem} with SIRIUS tool~\cite{duhrkop_sirius_2019} on MassSpecGym dataset, where MCES and Tanimoto metrics are calculated only on the valid SMILES that the method identified, which correspond to only 5.14\%.\\
$^*$ Baseline results for de novo molecule generation challenge on MassSpecGym, taken from \cite{bushuiev_massspecgym_2024}.\\ 
$\dagger$ Result taken from the respective indicated paper.\\
$\ddagger$ Results of baseline approaches implemented within DiffMS, taken from \cite{bohde_diffms_2025}, we assume all the models were evaluated on {NPLIB1-DiffMS} version of the dataset.
}
\label{tab:results-sota}
\end{table}
% \vspace{0.2cm}

\begin{figure*}[htb!]
\centering
\vspace{0.3cm}
\begin{center}
\includegraphics[width=\textwidth]{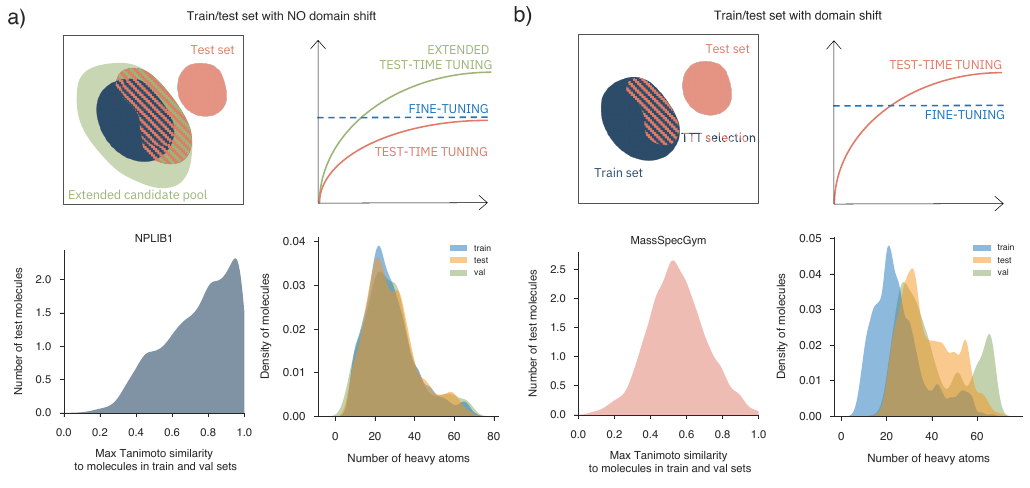}
\vspace{-0.5cm}
\caption{Overview of tuning strategies under varying domain conditions, accompanied by representative datasets.
(a) No domain shift: when the training and test sets share the same distribution, both fine-tuning and test-time tuning yield comparable performance. In such cases, fine-tuning typically represents the upper bound of achievable performance. However, performance can be further enhanced through extended test-time tuning, which leverages additional data (green) to expand the candidate pool. The NPLIB1 dataset exemplifies this scenario.
The second row illustrates the distribution of test molecules based on their maximum Tanimoto similarity to molecules in the training and test sets (left), as well as the distribution of molecules across the training, test, and validation sets based on their number of heavy atoms (non-hydrogen). The analysis is performed on {NPLIB1-Full}.
(b) Domain shift: in scenarios where the training and test sets differ significantly, fine-tuning may lead to degraded performance. In contrast, test-time tuning dynamically selects relevant samples, thereby improving generalization to the target distribution. The MassSpecGym benchmark serves as an example of this condition, with corresponding distributions shown in the second row. 
% Bottom row: a similar analysis is presented for the environmental application discussed in Section \ref{sec:env-app}. It shows the distribution of molecules in the complete dataset based on their number of heavy atoms (left), and their maximum Tanimoto similarity with the training, test, and validation sets of both NPLIB1 (blue) and MassSpecGym (red).
}
\vspace{-0.3cm}
\label{fig:analysis-datasets}
\end{center}
\end{figure*}

\subsection{In‑distribution regime: test‑time tuning achieves fine‑tuning performance
% Test-Time Tuning on small and informative datasets
}\label{sec:results-TTT-case1}
% NPLIB1 case
% Although uncommon, there are scenarios where moderately sized experimental datasets are available for training a machine learning model, and typically these datasets contain structures that are similar to each other or share similar features or properties. 
% Under such conditions, the training and test sets share the same underlying distribution, making fine-tuning on the training set highly effective for achieving optimal performance on the test set. 
We analyze here the case where the target spectra are generated by molecules that closely resemble those included in the training set. In this in-distribution setting, fine-tuning on the training set is highly effective for achieving optimal performance on the test set, since all the available data can be considered informative for the model to learn characteristics relevant to the test set. 
Consequently, standard fine-tuning serves as an upper bound for any improvements achievable through test-time tuning, as shown in Figure \ref{fig:analysis-datasets}a (top-right), where fine-tuning (blue) is represented as the asymptotic limit to test-time tuning (orange).\\
% explain more the figure, distributions etc...
A case study for this scenario is NPLIB1, where the test set is a hold-out from the same dataset, and the train set contains molecules similar to the ones in the test set. This is proven by Figure \ref{fig:analysis-datasets}a (bottom), where the distribution of the Tanimoto similarity of test molecules to the train molecules is skewed towards one, and the distribution of the number of heavy atoms in train, test and validation split overlap consistently.
We show in Table \ref{tab:results-TTT} that the fine-tuned model on {NPLIB1-Full} reaches Top--1 accuracy of 12.42\%, while the test-time tuned model reaches comparable performance at 12.21\% Top--1 accuracy. 
% As illustrated in Figure \ref{fig:ttt-set-sel-indices}, the algorithm selects almost all the candidate points during the training process, since they are all relevant for the test set. 
% The total amount of selected points converges to the size of the training set ($\sim$15 000) over the iterations.
% This confirms the fact that the test-time tuning framework is appropriate for in-distribution settings.\\
Interestingly, when test-time tuning is applied, the choice of starting from the pre-trained model or the fine-tuned one does not severely affect performance. By looking at the results in Table \ref{tab:results-TTT}, a relative gain of only 5\% is observed when starting from the fine-tuned model instead of from the pre-trained one, rendering prior fine-tuning negligible.\\
The upper bound imposed by standard fine-tuning can be meaningfully surpassed only by leveraging additional data, expanding the candidate pool with more experimental spectra, as shown in green Figure \ref{fig:analysis-datasets}a (top). 
In the case of NPLIB1-Full, the result is shown in the last (underlined) row of the top section of Table \ref{tab:results-TTT}, reaching a Top--1 accuracy of 17.28\%. 
% With this result we want to show that, given a set of unlabeled spectra, the proposed method is able to select the most relevant data from the candidate pool, obtaining higher performances if useful information is actually contained.
Refer to Section \ref{sec:results-ext-candidate-pool} for a more detailed description and analysis.

\subsection{Out‑of‑distribution regime: test‑time tuning enables domain adaptation
% Test-Time Tuning on large and heterogeneous datasets
}\label{sec:results-TTT-case2}
In situations where the acquired spectra originate from molecular structures that are poorly represented or entirely diverse from those in the training data, the problem shifts to an out‑of‑distribution regime. 
In out-of-distribution cases, the heterogeneity of the spectra used for training and across the observed molecules is so marked that some training points are only weakly informative for the target spectra being analyzed at test time. Consequently, fine‑tuning distracts the model during adaptation and leads to catastrophic forgetting of previously learned representations, thus degrading performance. 
Our results show that test‑time tuning is a robust alternative approach in these scenarios. \\
In MassSpecGym, train/test splits are constructed using the MCES distance, ensuring that no similar molecules are shared between train and test set. As Figure \ref{fig:analysis-datasets}b (bottom) illustrates, the train and test sets follow different distributions, making domain adaptation strategies essential to bridge the gap between source and target domains.
Fine-tuning on the entire MassSpecGym training set results in a performance drop compared to the pre-trained model, with Top--1 accuracy decreasing from 1.89\% to 1.13\% (c.f. Table \ref{tab:results-TTT}).
This indicates that using all available training data does not enhance learning; instead, it causes the model to forget knowledge acquired during pre-training. In contrast, applying test-time tuning yields a relative improvement of 67\%, reaching a Top--1 accuracy of 3.16\%, which is a notable improvement if compared to the baselines proposed in \cite{bushuiev_massspecgym_2024}. 
These results suggest that the training set does contain informative data points, but their effective use requires selective and adaptive strategies rather than broad fine-tuning. This highlights the importance of methods that can dynamically adapt to the target distribution without sacrificing previously learned knowledge.
% However, in this case the candidate pool is not extended to ensure that the splitting based on the MCES distance between molecules remains intact.

\vspace{0.5cm}
\begin{table}[ht]
    \centering
    \resizebox{0.9\textwidth}{!}{
    \begin{tabular}{lcccc}
    \toprule
    & Top--1 Accuracy $\uparrow$  & Top--5 Accuracy $\uparrow$  & Top--10 Accuracy $\uparrow$ & Valid SMILES $\uparrow$ \\
    \toprule 
    \multicolumn{5}{c}{\textsc{NPLIB1-Full}} \\
    \midrule
    Fine-tuning from scratch &  0.62\% & 1.98\% & 2.09\% & 11.86\% \\ 
    Zero-shot PT model & 3.84\% & 7.86\% & 9.17\% & 74.32\% \\ 
    {Fine-tuning PT model} & {12.42\%} & {25.91\%} & {30.83\%} & {88.01\%} \\
    Test-time tuning (from PT model)& {12.21\%} & {24.68\%} & {28.80\%} & {87.19\%} \\
    \textbf{Test-time tuning from FT model}& \textbf{12.88\%} & \textbf{26.38\%} & \textbf{31.32\%} & {89.12\%} \\
    Extended Fine-tuning (from PT model) & 7.16\% & 26.94\% & \underline{34.67\%} & \textbf{92.60\%} \\
    \underline{Extended Test-time tuning (from PT model)} & \underline{17.28\%} & \underline{27.72\%} & {31.74\%} & {84.68\%} \\
    \midrule
    \multicolumn{5}{c}{\textsc{MassSpecGym}} \\
    \midrule
    {Fine-tuning from scratch} & 0.00\% & 0.01\% & 0.02\% & 14.44\% \\
    Zero-shot PT model  & 1.89\% & 3.79\% & 4.39\% & {62.72\%} \\
    Fine-tuning (from PT model) & 1.13\% & 2.46\% & 2.93\% & \textbf{67.80\%} \\
    {Test-time tuning from FT model} & {1.26\%} & {2.60\%} & {3.22\%} & {67.76\%} \\
    \textbf{Test-time tuning (from PT model)} & \textbf{3.16\%} & \textbf{5.39\%} & \textbf{6.07\%} & {64.31\%} \\
    \bottomrule
    \end{tabular}
    }
    \vspace{0.2cm}
    \caption{Performances of the fine-tuned and test-time tuned models on the experimental datasets NPLIB1-Full and MassSpecGym. To highlight the impact of simulated data, the performances of the model trained from scratch on the experimental datasets are shown, as well as the zero-shot evaluation of the pre-trained model. \underline{Underlined} in the last row of the top section, we show the results of the test-time tuning strategy on NPLIB1, when the candidate pool is extended with additional experimental data (spectra from MassSpecGym).}
    \label{tab:results-TTT}
\end{table}
% \vspace{0.2cm}

% \subsection{Impact of simulations}
\subsection{Simulation-based pre‑training substantially enhances model performance}
\label{sec:results-impact-simulations}
Results demonstrate that our model can effectively learn from simulated spectroscopic data from \citet{alberts_unraveling_2024}, leading to marked improvements over training on experimental data alone. 
This highlights the value of simulation-based pre-training as a scalable solution to the limited availability of high-quality MS/MS experimental spectra, despite the observable differences between the two signals (see Figure \ref{fig:example-spectra1}). \\
%Their availability in large quantities makes them an attractive resource for both pre-training and downstream adaptation.
%This holds true also for MS/MS spectroscopy, where several techniques have been developed to simulate the fragmentation process of molecules.
% To quantify the improvements obtained by simulation-based pre-training, we compare the performance obtained by this strategy to the one obtained training the model from scratch only using the experimental data from NPLIB1 and MassSpecGym. 
%We demonstrate the considerable impact of simulated MS/MS spectra when used to pre-train our transformer model.\\
% , and then highlight the potential of incorporating simulations into the candidate pool from which the training data are selected during test-time tuning.\\
%We first assessed the model’s ability to learn directly from experimental data by training it 
%from scratch on NPLIB1 and MassSpecGym. 
Pre-training on simulated spectra leads to marked improvements in zero-shot performance, reaching 3.84\% Top--1 accuracy on NPLIB1-Full against 0.62\%, as shown by the first and second rows of Table \ref{tab:results-TTT}. 
%We pre-train our model on the dataset from \cite{alberts_unraveling_2024}, using spectra produced through multiple simulation modalities (see Section \ref{sec:methods-data}).
Performance on MassSpecGym also improved, rising from 0\% to 1.89\%. Although modest, the gain is noteworthy given the performances of almost all the current models being around 0\%~\citep{bushuiev_massspecgym_2024}, highlighting the difficulty of this benchmark.
In both cases, the fraction of valid SMILES among the Top--10 predictions by the model trained without simulations is very low, indicating that the experimental data alone from both datasets are insufficient for learning chemically consistent representations from scratch.
In contrast, adding simulated spectra in the pre-training stage provides a clear and substantial boost, addressing the limitations identified above.

% We cannot overlook the drop in performances when fine-tuning the model on MassSpecGym, where Top--1 accuracy fell to just 1.13\%. However, this can be attributed to the configuration of the dataset splits, as already discussed in Section \ref{sec:results-TTT-case2}.
%Despite the significant differences between simulated and experimental spectra (see Figure \ref{fig:example-spectra1}), we can conclude that during pre-training the model successfully learned transferable features linking MS/MS patterns to SMILES representations. Additionally, the percentage of valid SMILES predictions increased dramatically, indicating that pre-training also helped learning basic chemical rules.

\subsection{Leveraging additional experimental spectra improves test‑time adaptation on NPLIB1
% Extending test-time tuning on NPLIB1 with additional experimental spectra
}\label{sec:results-ext-candidate-pool}
We examine the effect of incorporating additional experimental spectra into the candidate pool of the test-time tuning of NPLIB1.
We expand the candidate pool of NPLIB1-Full by combining its original training set with MassSpecGym.
This extension leads to a substantial improvement in Top--1 accuracy, going from 12.21\% when using only the NPLIB1 training set, to 17.28\% after adding the extra spectra, equal to 41\% relative gain (cf. Table \ref{tab:results-TTT}). 
This demonstrates that the method is able to select training points containing relevant structural information for the test set, even from a large candidate pool of hundreds of thousand spectra.
% and this can be effectively leveraged by the tuning algorithm. 
% - MOVED TO DISCUSSION -
% while confirming that enlarging the candidate pool with experimental spectra can also significantly enhance performance, particularly in scenarios where the original training set is small. In practice, this suggests that curating and integrating additional experimental data should be a priority for improving model robustness and accuracy.

% \textcolor{blue}{plot number sim/exp points selected???}
\subsection{Improved molecular similarity and structural accuracy of predictions
% Correct molecular structure prediction
}\label{sec:results-similar-predictions}
\begin{figure*}[h]
\centering
\begin{center}
\includegraphics[width=\textwidth]{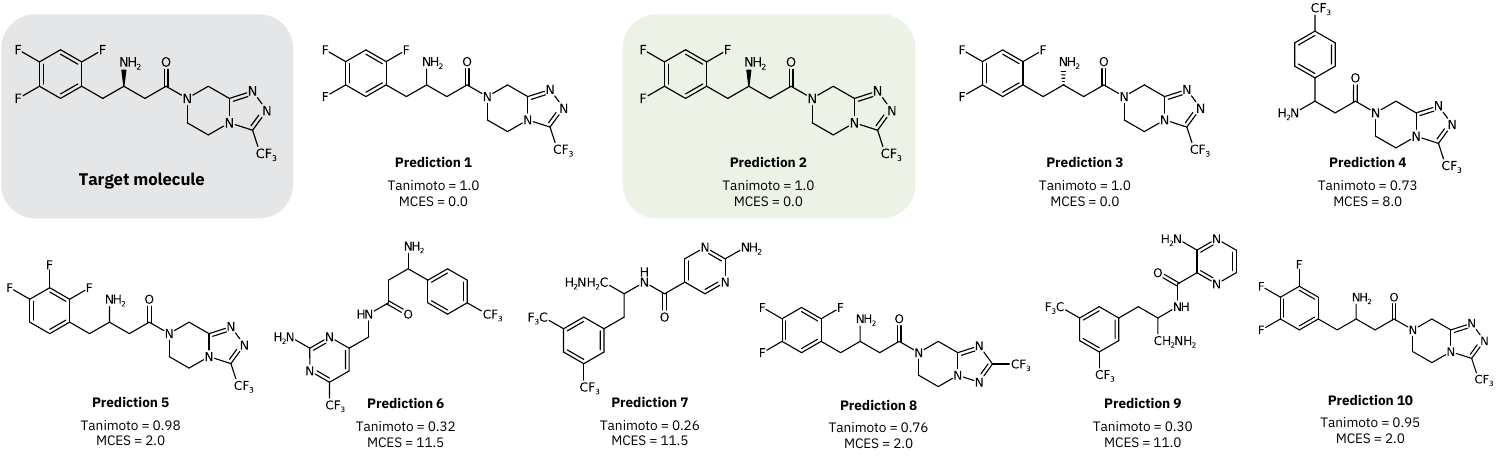}
\vspace{-0.3cm}
\caption{Top--10 predictions for one of the molecules present in the test set of MassSpecGym. Respective Tanimoto similarity and MCES distance from the target molecule are provided below every prediction. The model generates three stereoisomers among the first predictions, indicating structural awareness, but fails to identify the correct SMILES at the first attempt. The correct structure appears as the second candidate (highlighted in green), which positively contributes to the Top--10 accuracy.}
% \vspace{-0.3cm}
\label{fig:Top--10pred}
\end{center}
\end{figure*}

% In structure elucidation, the goal is of course to identify the molecule that generated the spectrum of interest, however, this is not always straightforward, due to the complexity of the spectrum and vastness of chemical space. So, even though an exact match is not directly found, whatever hint on the molecular structure can considerably reduce the number of candidate molecules we have to consider to find the exact correct molecule. 
% Consequently, a model that is able to predict meaningfully similar molecules to the target one is greatly useful.
Even when the model fails to predict the correct molecular structure, its predictions still provide relevant 
%A model capable of proposing molecules that are meaningfully similar to the true structure is highly valuable in structure elucidation. Since spectra are often complex and chemical space is vast, identifying the exact generating molecule is not straightforward, so that even without an exact match, 
guidance toward structurally relevant candidates, which can aid chemists.
The Tanimoto similarity and MCES distance can be seen as proxies for how two molecules are closely related to each other, and hence how informative a predicted structure may be to a chemist when narrowing down the search space for the true structure. 
The higher the similarity, the closer the predicted structure is to the actual target, and, inversely, the smaller the distance, the easier it should be to converge to the true structure. \\
%We evaluate the ability of our approach to predict molecules that are closely related to the target in terms of two popular measures of distance between molecules, namely the Tanimoto similarity and MCES.\\
% To do so, Tanimoto similarity and MCES distance~\cite{mces} are used.\\
% Tanimoto similarity is the most common computed metric, measuring the overlap of structural features between two different molecules. It ranges from 0, if there is no overlap, to 1, if the two molecules are exactly the same. In general, two molecules with Tanimoto similarity $> 0.85$ are considered very similar.
% MCES distance measures instead how different two molecules are by relying on the largest common edge subgraph between the two, focusing on shared chemical bonds \cite{mces}. If two molecules are the same, their MCES distance is 0 and it increases with them being more and more different.\\
% In this section we show more in details how our model predicts molecules that are in general very similar to the target.
For all the datasets in Table \ref{tab:results-sota}, the Tanimoto similarity is higher for the predictions of our framework than those of other models. 
The average Tanimoto similarities of the Top--1 predicted molecules (excluding the invalid SMILES) are 0.62 on NPLIB1-Full and 0.46 on MassSpecGym, hence 
% making them 
meaningful matches on average. 
Similarly, the MCES distance is lower for our approach than the baselines, with MCES reaching 6.28 on NPLIB1-Full and 11.77 on MassSpecGym.
Results are similar on NPLIB1-DiffMS and MassSpecGym-DiffMS, for which both metrics improve even further (c.f. Table \ref{tab:results-similarity}).\\
Taking inspiration from the analysis done in \citet{bohde_diffms_2025}, we show Table \ref{tab:results-similarity}, where we classified the molecules depending on their Tanimoto similarity with the respective target. More precisely, two classes are introduced: a meaningful match is defined if Tanimoto similarity $\geq 0.4$, while a close match in case Tanimoto similarity $\geq 0.675$.
While meaningful matches indicate general structural correctness, close matches reflect near-identical chemical similarity, which is more challenging to achieve. 
The proposed approach outperforms existing methods in both classes, 
% its performance on close matches is particularly noteworthy: for NPLIB1-DiffMS, it achieves around 44\% Top--1 when applying test-time tuning, and 48.13\% when fine-tuning, far surpassing other methods in both cases. 
% Even on the more complex MassSpecGym, it maintains the lead with almost 10\% of close matches for the Top--1 predictions on the full dataset and 9.46\% on MassSpecGym-DiffMS. 
achieving nearly 350\% increase in Top--1 meaningful matches (56.22\%) compared to DiffMS (12.41\%). 
These results underscore the model’s strength in generating highly accurate structures, not just broadly correct ones.\\
% The majority of all the predicted SMILES is valid, mainly thanks to the pre-training on simulated data, as already discussed (see Section \ref{sec:results-impact-simulations}). However, besides achieving great results in all the other metrics, our model predicts a lower percentage of valid SMILES compared to the other methods. \\
To conclude, Figure \ref{fig:Top--10pred} presents an example of predicted molecules for a given target. Although the model does not identify the exact molecule on its first attempt, it succeeds on the second. Interestingly, the first three predictions are stereoisomers with a Tanimoto similarity of 1.0, highlighting the model’s strong understanding of the target structure.

\vspace{0.2cm}

\begin{table}[ht]
\vspace{0.5cm}
\centering
\resizebox{\textwidth}{!}{%
    \begin{tabular}{lccccc}
    \toprule 
    \multirow[c]{3}{*}{MODEL}& \multicolumn{2}{c}{Top--1} & \multicolumn{2}{c}{Top--10} & \multirow[c]{3}{*}{Valid SMILES $\uparrow$}  \\
    \cmidrule(l{0.15cm}r{0.15cm}){2-3}
    \cmidrule(l{0.15cm}r{0.15cm}){4-5}
     &  Meaningful match $\uparrow$ & Close match $\uparrow$ & Meaningful match $\uparrow$ & Close match $\uparrow$ & \\
     &  ($\geq 0.4$) & ($\geq 0.675$) & ($\geq 0.4$) & ($\geq 0.675$)& \\
    % \midrule
    \toprule 
    \multicolumn{6}{c}{\textsc{NPLIB1-Full}} \\
    \midrule
    {This work (FT)} & {72.01\%} & {43.14\%}& \textbf{89.00\%}& \textbf{62.41\%}& {88.01\%}\\
    {This work (TTT)} & 70.12\% & {38.71\%}& {88.34\%}& {58.94\%}& {87.19\%}\\
    \textbf{This work (TTT-FT)} & \textbf{72.51\%} & \textbf{43.81\%}& {88.62\%}& {62.38\%}& \textbf{89.11\%}\\
    % \textcolor{red}{This work (Extended TTT)} & \% & \textbf{\%}& \textbf{\%}& \textbf{\%}& \textbf{\%}\\
    \midrule
    \multicolumn{6}{c}{\textsc{NPLIB1-DiffMS}} \\
    \midrule
    Spec2Mol~\cite{spec2mol}$\ddagger$ & 0.00\% & 0.00\% & 0.00\% & 0.00\% & 66.5\% \\
    MIST + Neuraldecipher~\cite{goldman_mist-cf_2024,le_neuraldecipher_2020}$\ddagger$ & 29.30\% & 7.33\% & 41.39\% & 12.82\%& 91.11\% \\
    MIST + MSNovelist~\cite{goldman_mist-cf_2024,stravs_msnovelist_2022}$\ddagger$ & 32.90 \% & 11.78\% & 44.79\% & 19.02\%& 98.60\%  \\
    DIFFMS~\cite{bohde_diffms_2025}$\ddagger$ & 27.40\% & 12.83\% & 46.45\% & 22.04 \%& \textbf{100.0\%}  \\
    {This work (FT)} & \textbf{75.00\%} & \textbf{48.13\%}& {90.11\%}& {65.37\%}& {90.24\%}\\
    \textbf{This work (TTT)} & 72.60\% & {43.99\%}& \textbf{91.50\%}& \textbf{65.45\%}& {89.28\%}\\
    % \textcolor{red}{This work (Extended TTT)} & \% & \textbf{\%}& \textbf{\%}& \textbf{\%}& \textbf{\%}\\
    \midrule 
    \multicolumn{6}{c}{\textsc{MassSpecGym}} \\
    \midrule
    \textbf{This work (TTT)} & \textbf{55.53\%} & \textbf{9.99\%}& \textbf{76.07\%}& \textbf{17.03\%}& \textbf{64.31\%}\\
    \midrule 
    \multicolumn{6}{c}{\textsc{MassSpecGym-DiffMS}} \\
    \midrule
    Spec2Mol~\cite{spec2mol}$\ddagger$ & 0.0\% & 0.0\% & 0.0\% & 0.0\%& 68.5\% \\
    MIST + Neuraldecipher~\cite{goldman_mist-cf_2024,le_neuraldecipher_2020}$\ddagger$ & 0.29\% & 0.01\% & 0.39\% & 0.09\%& 81.78\% \\
    MIST + MSNovelist~\cite{goldman_mist-cf_2024,stravs_msnovelist_2022}$\ddagger$ & 0.66\% & 0.00\% & 1.92\% & 0.00\%& 98.58\% \\
    DIFFMS~\cite{bohde_diffms_2025}$\ddagger$ & 12.41\% & 3.78\% & 32.47\% & 6.73\%& \textbf{100.0\%} \\
    \textbf{This work (TTT)} & \textbf{56.22\%} & \textbf{9.46\%}& \textbf{76.17\%}& \textbf{16.93\%}& {64.11\%}\\
    \bottomrule
    \end{tabular}
}
\caption{Additional evaluation of the similarity of predicted molecules to the target molecule. Results for our models and comparison to others are provided depending on the different datasets. Different classes are defined depending on the Tanimoto similarity, as reported at the top of the table and explained in Section \ref{sec:results-similar-predictions}. Percentage of valid SMILES on the first 10 predicted molecules for our models, and over all the predictions for the others (see Appendix, Table 4 in \citet{bohde_diffms_2025}).
Highlighted in bold the name of the model holding the best Top--1 accuracy on the left, while the best score in each class on the right.\\
$\ddagger$ Results of baseline approaches implemented within DiffMS, taken from \citet{bohde_diffms_2025}, we assume all the models were evaluated on the same NPLIB1 filtered version that we named {NPLIB1-DiffMS}.}
\vspace{0.7cm}
\label{tab:results-similarity}
\end{table}

%% file: sections/discussion.tex
We demonstrate that transformer-based language models, combined with test-time tuning, offer a promising direction for advancing de novo molecular structure elucidation from MS/MS spectra. Test-time tuning has been especially impactful in mitigating out-of-distribution scenarios. 
By eliminating intermediate steps such as fragment annotation, our approach achieves true end-to-end generation of SMILES strings from spectra and chemical formulae. 
This design not only simplifies the pipeline but also improves the understanding of chemical structures, as evidenced by the high structural similarity of predicted candidates even when exact matches are not obtained.\\
Our framework proves powerful across different scenarios. 
In domains with minimal distribution shift, such as NPLIB1, test-time tuning achieves comparable results to classical fine-tuning, demonstrating its effectiveness even when adaptation is less critical.
% while even higher performances can be reached when additional informative data are available. 
In contrast, in highly heterogeneous settings where domain shift is pronounced, like in MassSpecGym case, test-time tuning becomes essential to obtain improvements in performance otherwise lost with naive broad adaptation. 
This flexibility highlights the robustness of our approach and its ability to adapt dynamically to diverse data conditions without sacrificing previously learned knowledge.\\
% add here comment on additional data
Moreover, the inclusion of more experimental spectra can provide richer structural information, potentially enabling the model to learn more effectively and significantly enhance performance, particularly in scenarios where the original training set is small, as in NPLIB1 case. 
This suggests that curating and integrating additional experimental data should be a priority for improving model robustness and accuracy.\\
% Our experiments also indicate that having access to a larger pool of experimental data, increases the likelihood of selecting informative samples during adaptation, which in turn improves generalization and chemical plausibility of predictions. \\
The impact of simulated data is particularly noteworthy. 
Pre-training on large-scale simulations enhances zero-shot performance 
but also provides the model with a richer understanding of fragmentation patterns and structural relationships, enabling it to generalize better to unseen molecules.
This foundational knowledge significantly reduces the limitations imposed by scarce experimental data and sets the stage for more effective fine-tuning and adaptation.\\
% While simulations cannot fully replace real spectra, they offer a scalable and complementary resource that mitigates data scarcity and improves resilience to domain shifts.\\
Finally, our evaluation shows that even when the predicted SMILES does not perfectly match the ground truth, the generated candidates remain chemically meaningful. High Tanimoto similarity and low MCES distances indicate that these predictions provide valuable structural hints, significantly narrowing the search space for human experts. This property transforms the model from a mere predictor into a practical assistant for structure elucidation, offering actionable insights rather than isolated guesses.

% limitations
While the proposed framework demonstrates strong performance across datasets and distributional regimes, it also presents a few practical limitations. The primary challenge concerns scalability: as the candidate pool grows, identifying the most informative spectra for test‑time tuning becomes increasingly demanding. Approximate nearest-neighbor or vector‑search methods could help mitigate this bottleneck, but integrating such retrieval steps without degrading adaptation quality remains non‑trivial.\\
A more fundamental limitation is the method’s dependence on having at least some structurally relevant spectra within the candidate pool. In regions of chemical space where informative neighbors are absent, the ability of test‑time tuning to improve predictions is inherently constrained. Moreover, the quality of the spectra themselves plays a critical role: noisy or low-quality spectra can obscure structural signals, reducing the effectiveness of both the selection process and subsequent model updates. This dependency underscores the method’s core assumption: test‑time tuning is highly effective when relevant and reliable information exists, but its benefits diminish when the available data contain no actionable cues.\\
Nevertheless, an important advantage of test‑time tuning lies in its efficiency: unlike conventional fine‑tuning, it requires substantially fewer data points for adaptation. When the selection of informative spectra can be performed rapidly, this property translates into faster overall adaptation without sacrificing predictive performance, making it particularly appealing in scenarios where computational resources or time are limited.

As the first application of test‑time tuning to structure elucidation from spectroscopic data to the best of our knowledge, the results highlight the potential of adaptive language models to transform MS/MS-based structure elucidation workflows. By leveraging test-time tuning and simulated data, these models offer a scalable and flexible solution for navigating chemical diversity, paving the way for more accurate and efficient identification of unknown compounds in metabolomics, natural product discovery, and beyond.

%% file: sections/methods.tex
\subsection{Datasets}
\begin{figure}[ht]
    \includegraphics[width=\textwidth]{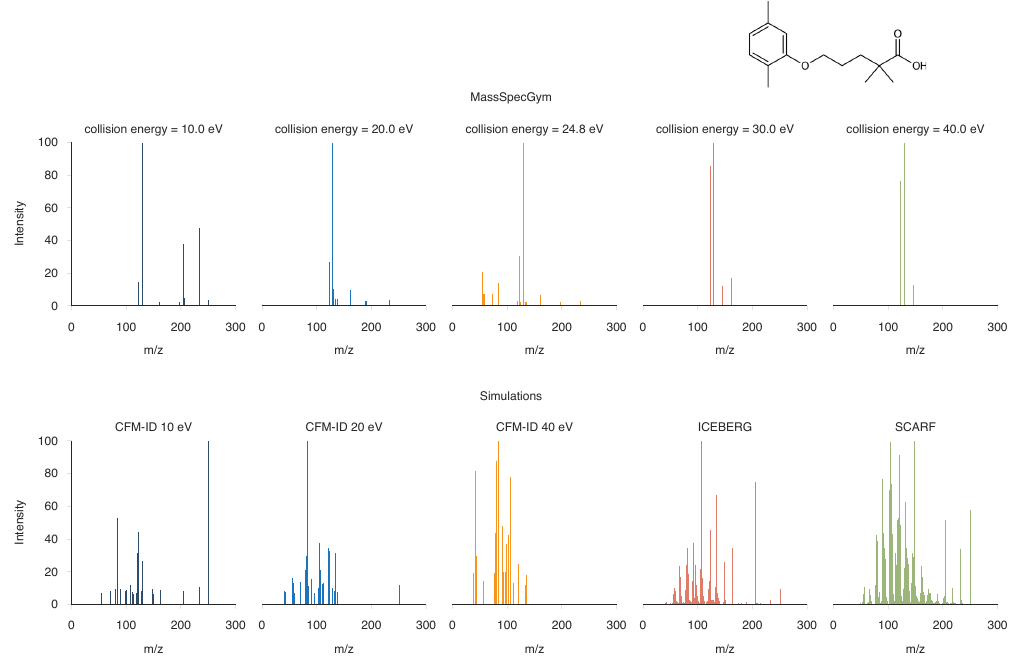}
    \vspace{-0.3cm}
    \caption{Example of available spectra for one specific molecule in the  datasets used in the present work. In this case MassSpecGym~\cite{bushuiev_massspecgym_2024} contains 5 different spectra obtained using different values for the collision energy. In the last row are presented the spectra obtained with the 5 simulation techniques used in ~\cite{alberts_unraveling_2024} and mentioned above. As it is possible to see they all severely differ from each other.}
    \label{fig:example-spectra1}
\end{figure}

We use three different datasets of positive mode MS/MS spectra relying on H$^+$ and Na$^+$ adducts.
Simulated MS/MS spectra are obtained from~\cite{alberts_unraveling_2024}, which sums up to a total of 3 971 930 simulations combining CFM-ID 4.0 \cite{wang_cfm-id_2021} with collision energy equal to $10\ eV$, $20\ eV$ and $40\ eV$, ICEBERG \cite{goldman_generating_2024}, and SCARF \cite{goldman_prefix-tree_2023}.
An example of the spectra obtained with these techniques is shown in Figure \ref{fig:example-spectra1}.
As for the experimental spectra, we used NPLIB1, which is derived from GNPS~\cite{wang_sharing_2016} and was first introduced in \cite{duhrkop_systematic_2021}, and MassSpecGym~\cite{bushuiev_massspecgym_2024}, which already provides a fixed train, validation and test split to benchmark against. These datasets contain respectively 19 687 and 231 104 spectra. \\
However, to allow a fair comparison with results from other methods, we also evaluated different versions of the same datasets. More specifically, the pre-processing applied in \citet{bohde_diffms_2025} is performed, consisting in an adduct-based and element-based filtering operation. Only spectra containing H$^+$ adducts, and molecules containing only carbon, oxygen, nitrogen, hydrogen, phosphorus, sulfur, chlorine and fluorine are retained. 
We call these versions of the datasets {MassSpecGym-DiffMS}, with 227 341 spectra, and {NPLIB1-DiffMS} with 7 947 spectra. More details can be found in Appendix \ref{sec:appendix-data}.\\
%The MassSpecGym dataset is already split between train, validation and test set. This was performed by taking into account the distance between the molecules in different sets via the Maximum Common Edge Subgraph (MCES) distance, so that the test set can be considered out-of-distribution with respect to the train set \cite{bushuiev_massspecgym_2024}, making this dataset extremely challenging.
When looking at the spectra obtained for a specific molecule, the simulated ones significantly differ from the ones obtained through experiments, as can be seen in Figure \ref{fig:example-spectra1}, which underscores the challenge of bridging the gap between simulated and experimental spectra during model training and adaptation.
% In the end, we show the utility of our method by evaluating it on a real-world application. The dataset we use contains 445 MS/MS spectra of commercially used antioxidants and a small number of their known transformation products, it was collected and provided by Eawag. More details can be found in \ref{sec:appendix-data}.\\

% Data in short, example of spectra

\subsection{Architecture and modalities}\label{sec:methods-models}
The proposed model follows a sequence-to-sequence encoder-decoder architecture. It takes as input the MS/MS spectrum and the chemical formula of a molecule to predict the corresponding SMILES, as illustrated in Figure \ref{fig:schema_TTT}. 
We decided to include the chemical formula in the inputs of the model since it is usually known when tandem mass spectroscopy is performed either computationally or during experiments. 
Its inclusion allows the model to gather more information about the elements present in the target molecule.
Every modality, meaning spectrum, chemical formula and SMILES, is treated as text. In particular, the peaks of each spectrum are encoded as a list of tuples of the mass-to-charge ratio and intensity $[mzs, I]$ and converted then to a string.
%, as shown in Figure \ref{fig:tokenization}.
%Multimodal embedding is implemented to combine the different types of data, with application of sinusoidal positional encoding. 
Further details can be found in Appendix \ref{sec:appendix-model}.

\subsection{Fingerprint alignment}
An additional component of the model is a multilayer perceptron (MLP) placed on top of the encoder, which predicts molecular fingerprints from the encoder’s output embeddings (see Figure \ref{fig:schema_TTT}). To account for this prediction task, a binary cross-entropy loss term is incorporated alongside the standard cross-entropy loss used for the encoder–decoder. This, often referred to as fingerprint alignment, allows the model to learn representations that better capture chemical information, guiding SMILES predictions toward more plausible structures (see Equation \ref{eq:loss} in Appendix \ref{sec:appendix-model-fing-alignment}).

\subsection{Formula constrained generation}
To further leverage the chemical formula provided as input, we implement formula-constrained generation at prediction time~\cite{alberts_setting_2025}. Since the model generates SMILES strings token by token, we dynamically restrict the set of allowed tokens at each decoding step by removing those that would violate the given chemical formula. This ensures that the generated SMILES is always consistent with the specified formula, thereby increasing the likelihood of producing the correct target structure. Beyond improving validity, this constraint also guides the model toward chemically plausible candidates, enhancing both accuracy and interpretability.

\subsection{Rejection sampling}
Alongside formula constraints, we apply rejection sampling to better exploit the model’s ability to generate meaningful candidates, while reducing the impact of invalid SMILES predictions. 
Specifically, the model generates 50 SMILES strings, and we retain the first 10 (or less if there are not enough) that pass chemical syntax and structural validity checks.

\subsection{Test-time tuning}\label{sec:methods-ttt}
Test-time tuning is a transductive strategy~\citep{book_transduction,farahani_brief_2021,sun2020test} that updates parameters at inference time using only unlabeled test inputs, contrasting with pre‑training/fine‑tuning schemes that assume target data are available during training, to better align predictions with the characteristics of the input data. Unlike conventional training, which relies on a fixed dataset, this method exploits information available at inference time to refine the model without requiring full retraining. The process typically involves selecting relevant training points from a large candidate pool, often using a nearest-neighbor strategy (see Figure \ref{fig:schema_TTT}). 
Test-time tuning is particularly valuable in scenarios involving domain shifts between training and test sets~\citep{hubotterefficiently}, where labeled data exist only in the source domain but some unlabeled target data can be leveraged during inference —a setting commonly referred to as transductive transfer learning. By dynamically adapting to the test distribution, this strategy improves robustness and predictive accuracy, especially in challenging conditions where distribution shifts would otherwise degrade performance.

\subsection{Selection of relevant data points}
\begin{figure*}[htb!]
\centering
\begin{center}
\includegraphics[width=0.49
\textwidth]{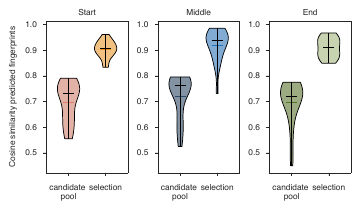}
\includegraphics[width=0.49
\textwidth]{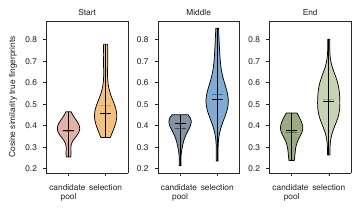}
\caption{Comparison of the cosine similarity distribution of the predicted (left) and true (right) fingerprints of the candidate pool and selected portion with the considered test point, over different iterations at the beginning, during and at the end of the tuning process. }
\label{fig:ttt-selection-distribution}
\end{center}
\end{figure*}

In our implementation, 
we first predict molecular fingerprints for test spectra using an auxiliary MLP head on the transformer encoder. We then perform k‑means clustering over test‑set fingerprints to uniformly sample the test space. Iterating over clusters, we select one representative test instance per cluster (closest to the centroid) and retrieve a batch of candidate training points from the pool based on cosine similarity in fingerprint‑logit space. 
% test-time tuning is applied to remedy the domain shift that is present in certain cases between the train and test set of MS/MS experimental datasets, helping the model to predict more accurate molecular structures. 
% Usually the candidate pool of data points matches the training set of the given dataset, however, it is always possible to add more data to it, which results extremely useful when other relevant data to the test set are available, as shown in Section \ref{sec:results-ext-candidate-pool}.\\ 
% To achieve a more uniform sampling of the test set space without having to use every test point, we first apply k-means clustering to the predicted fingerprints of the test set. 
% The algorithm then iterates over the clusters and selects the closest test point to the respective centroid and uses it to select a new batch of training samples from the candidate pool. 
These training points correspond (if the representation is good enough) to the most relevant data points for the current test instance. 
The selection is typically done through a nearest-neighbor search in the embedding space.
Selected samples are used for one‑step gradient updates to the encoder–decoder and fingerprint head; embeddings (and fingerprints) are periodically refreshed to reflect updated parameters.
% When the candidate pool is very large, computing embeddings and fingerprints becomes computationally expensive in terms of time and memory, which can significantly degrade performance. To address this, 
% we adopt 
% a pre-selection strategy
% can be adopted. Using the FAISS library~\cite{faiss}, we perform k-means clustering on the candidate data points and select the cluster whose centroid is closest to the test set. This smaller subset serves as the candidate pool for the next 10 iterations. After every 10 parameter updates, the clustering and selection process is repeated to ensure that the chosen subset remains relevant to the test set, as the model’s representations evolve during tuning.
In Figure \ref{fig:ttt-selection-distribution}, we show how, already from the first iterations, the method selects data points from the candidate pool that have fingerprints with cosine similarity closer to the considered test point than the rest of the candidate pool. In addition, the average cosine similarity of the selected points to the test point slightly increases over the iterations.

%% file: sections/appendix.tex
% APPENDIX
\appendix\label{sec:appendix}
\onecolumn

\section*{Appendix}
\section{Data}\label{sec:appendix-data}
This work is limited to tandem spectra generated via positive electrospray technique.
When the splitting is not provided, we divide the data approximately as: 20\% test set, 5\% validation set and 75\% training set, always ensuring different spectra obtained for the same compound falls in the same split (to avoid data leakage).

The pre-training of the model was performed using the dataset of simulated MS/MS spectra from \citet{alberts_unraveling_2024}. It contains 794 386 entries, each describing a SMILES-MS/MS spectra tuple. Every entry indeed contains 5 different spectra obtained using different simulation modalities, namely CFM-ID 4.0 \cite{wang_cfm-id_2021} with collision energy equal to $10\ eV$, $20\ eV$ and $40\ eV$, ICEBERG \cite{goldman_generating_2024} and SCARF \cite{goldman_prefix-tree_2023}. As shown in Figure \ref{fig:example-spectra1}. In total, the dataset contains 3 971 930 spectra.
The count of unique SMILES is 789 328.
The spectra have been normalized to have maximum intensity equal to 100, afterwards, all the peaks with intensity smaller than 1 have been removed, since considered noise. After this pre-processing step every spectrum contains at most 31 peaks for the three CFM-ID modalities, 300 for the SCARF modality, while 595 for ICEBERG. 

The tuning and evaluation of our framework was performed using two different experimental datasets, MassSpecGym and NPLIB1. In particular, different versions of each dataset have been introduced in other works, we evaluate our model on several of these, to ensure a fair comparison with the already publicly available results. Here we provide a detailed description of all the versions.  
Respective links or pointers to the material to easily obtain and format each dataset are given in the text below. 

\paragraph{MassSpecGym}
To the best of our knowledge, it is the most challenging dataset for MS/MS data, obtained from \citet{bushuiev_massspecgym_2024}, it contains 231 104 spectra for 31 602 unique compounds (after canonicalization). Different instruments were used for the collection of these spectra, such as varying collision energy and H$^+$ and Na$^+$ adducts.
Train, validation and test splits are given, since they were created using a threshold of 10 on the MCES distance between the molecules in train, validation and test set, which is what makes the dataset so challenging. In particular, the test set contains 17 556 spectra. It can be downloaded from \url{https://huggingface.co/datasets/roman-bushuiev/MassSpecGym/blob/main/data/MassSpecGym.tsv}.
To avoid confusion, we refer to this original version of the dataset as just \textsc{MassSpecGym}. \\
In fact, another version of the dataset is introduced in \citet{bohde_diffms_2025}, where a filtering on the elements contained in the molecules is performed. More precisely, only compounds with carbon, oxygen, nitrogen, hydrogen, phosphorus, sulfur, chlorine and fluorine elements were kept, for a total of 30 640 unique compounds, 227 341 spectra, of which only 17 082 in the test set (instead of the initial 17 556 spectra). See Table \ref{tab:info-datasets} for more details.
We refer to this filtered version of MassSpecGym as \textsc{MassSpecGym-DiffMS}.

\paragraph{NPLIB1} The second experimental dataset we used is NPLIB1, which was derived from GNPS library~\cite{wang_sharing_2016} and MassBank~\cite{massbank}, and firstly introduced in \cite{duhrkop_systematic_2021}, where it was used for the SVM training (and not to be confused with the structures dataset instead used for CANOPUS training\footnote{For full clarity, a list of spectra files is also available in Supplementary Table 6 of the respective manuscript ~\cite{duhrkop_systematic_2021}, which shows 10 710 file names. However, it refers to a slightly different dataset, which was instead used for the training of CANOPUS tool, instead of the SVM presented in the manuscript. Indeed, 4 file names in this list are not present in the SVM training dataset (ids: CCMSLIB00000078787, CCMSLIB00000081065, CCMSLIB00000847829, CCMSLIB00000855420), while 3 different file names were added (ids: CCMSLIB00001058585, CCMSLIB00001058867, CCMSLIB00001059041).}). 
It can be downloaded from \url{https://bio.informatik.uni-jena.de/wp-content/uploads/2020/08/svm_training_data.zip}, while a full description of the data they used is available at \url{https://bio.informatik.uni-jena.de/data/}. The folder contains 10 709 spectra files (see `compound\_ids.tsv` file in the downloaded folder) for a total of 19 687 spectra, since multiple spectra with different collision energies were measured for several compounds, and 8553 unique compounds. The spectra were obtained using both H$^+$ and Na$^+$ adducts. More details can be found in Table \ref{tab:info-datasets}.
We refer to this dataset as \textsc{NPLIB1-Full}.\\
This dataset has been used in later works, such as \cite{goldman_annotating_2023, goldman_mist-cf_2024, goldman_prefix-tree_2023} and \cite{bohde_diffms_2025}, where it was subject to filtering procedures, generating several different versions of the dataset.
First of all, MIST tool~\cite{goldman_annotating_2023}, which can be used to annotate tandem mass spectra peaks with chemical structures, was trained on a dataset composed by both commercially and publicly available data. The public dataset is composed by the NPLIB1 spectra obtained by using only H$^+$ adduct. 
After this adduct-based filtering, only 8030 spectra files remain, which contain a total of 14 532 spectra with different collision energies, corresponding to 7131 unique compounds\footnote{The initial folder containing all the 10 709 spectra files can also be downloaded from \url{https://zenodo.org/records/8316682}.}. The same dataset was then also used in \citet{goldman_mist-cf_2024}. However, all these works mainly addressed spectra predictions tasks instead of structure elucidation, as done in this work, and they processed the dataset by merging all the peaks present in a single spectra file, and obtained under different collision energies, in only one spectra.
On the other hand, for our purpose, it is straightforward to separate the peaks obtained at different collision energies in different spectra objects, ending up with multiple spectra for several files. 
We refer to this dataset as \textsc{NPLIB1-H$^+$}.\\
In the end, also DiffMS~\cite{bohde_diffms_2025} was evaluated on the structure elucidation task for NPLIB1 dataset, however, as for MassSpecGym dataset, they performed an additional filtering on top on the adduct-based one, so that only spectra of molecules containing only carbon, oxygen, nitrogen, hydrogen, phosphorus, sulfur, chlorine and fluorine elements were kept. Resulting with a total of 7947 files, containing 7127 unique compounds.
Moreover, they did not merge all the peaks in each file in a unique spectrum as mentioned before for other works, but most likely used only the first spectrum written in every file, ending up with the same number of spectra as the files, equal to 7947, with 6748 in training set, 396 in validation set and 803 in test set. 
We obtained this dataset by running DiffMS code with the configurations and parameters they provided to reproduce the results. In particular, the train-val-test split used can be found in the file `DiffMS/data/canopus/splits/canopus\_hplus\_100\_0.tsv`.
We additionally evaluate our method on this last version of the NPLIB1 dataset to fairly compare with DiffMS results, and we refer to it as \textsc{NPLIB1-DiffMS}.

% \paragraph{Antioxidants application} This dataset was provided by Eawag, who collected and curated the data following their internal standard operating procedures for measuring spectra of chemical standards for upload to MassBank \cite{massbank}. All sampling, measurement, and initial processing steps were performed by Eawag; our contribution focuses exclusively on computational analysis and modeling. 
% It contains 445 files, each one containing a single spectra measured in positive ionization mode. 
% We never train the model on this dataset, only evaluation is performed using different models, either fine-tuned on the other experimental datasets, or test-time tuning is performed using this dataset as test set, while the model is trained only on the data present in the candidate pool (See Section \ref{sec:methods-ttt}).

Notebooks for the construction of all the different versions of the datasets are available in the public code of this work.

We did not discard any spectrum from any dataset, only normalization was applied and noisy peaks removed. More precisely, for every spectrum, the peaks were normalized such that the maximum intensity was equal to 100, consequently, the peaks with intensity smaller than 1 were removed. All the SMILES strings were canonicalized and Morgan fingerprint with size 128 calculated using rdkit \cite{rdkit}.

% short analysis datasets
\begin{table}[ht]
    \centering
    \resizebox{\textwidth}{!}{%
    \begin{tabular}{lccccccc}
    \toprule
     & Simulations & NPLIB1-Full & NPLIB1-H$^+$ & NPLIB1-DiffMS & MassSpecGym & MassSpecGym-DiffMS \\
    \toprule 
     \# files & - & 10709 & 8030 & 7947 & - & - \\
     \# spectra & 3971930 & 19687 & 14532 & 7947 & 231104 & 227341 \\
     \# unique SMILES & 789328 & 8553 & 7131 & 7053 & 31602 & 30640\\
     Split provided$^*$ & \xmark & \xmark & \xmark & \checkmark & \checkmark & \checkmark \\
     \# train set & 2978947 & 14897 & 10927 & 6748 & 194119 & 191216 \\
     \# val set & 198597 & 908 & 622 & 396 & 19429 & 19043 \\
     \# test set & 794386 & 3882 & 2983 & 803 & 17556 & 17082 \\
     \# unique SMILES in test set & - & 1711 & 1427 & 701 & 3170 & 3076 \\
     Adducts & H$^+$ & H$^+$, Na$^+$ & H$^+$& H$^+$ & H$^+$, Na$^+$ & H$^+$\\
     % min(\textit{mzs}) & 1.00 & 2.39 &  & & 2.39 \\
     % max(\textit{mzs}) & 1084.22 & 2005.47 &  & & 2881.13 \\
     % min(\# peaks) & 1 & 1 & 1 & 1 & 1 & 1\\
     % max(\# peaks) & 595 & 49177 & 48499 & 2104 & 299 & 299\\
     C,H,O,P,N,S,Cl,F & \checkmark & \checkmark & \checkmark & \checkmark & \checkmark & \checkmark \\
     B,Br,I & \checkmark & \checkmark & \checkmark & \xmark & \checkmark & \xmark\\
     Si & \checkmark & \checkmark & \checkmark  & \xmark & \checkmark & \xmark\\
     Se & \xmark & \checkmark & \checkmark & \xmark & \checkmark & \xmark\\
     As & \xmark & \xmark & \xmark & \xmark & \checkmark & \xmark \\
     Sn,Al & \xmark & \xmark  & \xmark & \xmark & \xmark & \xmark \\
    \bottomrule
    \end{tabular}
    }
    \caption{Details about the content of the different datasets used. In particular, the number of spectra files used, the effective total number of spectra (by dividing peaks obtained at different collision energies in different spectra files), number of unique compounds. We also report if the train-val-test split was given, together with the number of spectra in each set. We also checked the elements present in each datasets version.\\
    $^*$ When the train-val-test split is not provided, we perform a 75-5-20 split ensuring no spectra for the same compound fall in different sets. If provided, we adopted the given one and simply report the numbers.\\}
    \label{tab:info-datasets}
\end{table}
\vspace{1cm}

\section{Model and training details}\label{sec:appendix-model}
We train a transformer encoder–decoder with $\sim$150 million parameters using cross-entropy loss for SMILES generation and an additional binary cross-entropy loss for fingerprint prediction. 
The model is configured with a hidden dimensionality of 1024 and multi-head attention with 8 heads for both the encoder and decoder. The architecture consists of 6 layers in the encoder and 6 layers in the decoder, each with a feed-forward network dimension of 2048. Normalization strategies include multimodal normalization, final layer normalization, and post-layer normalization, while gated linear units are enabled to improve representational capacity. 
Optimization is performed using the AdamW algorithm~\cite{adam} with $\beta_1$=0.9 and $\beta_2$=0.999, and no weight decay is applied. 
The exponential learning rate scheduler is used with initial learning rate of $1e^{-4}$ for pre-training and $5e^{-5}$ for fine-tuning and test-time tuning, with a decay factor $\gamma$ of 0.95 for standard pre-training and fine-tuning, and 0.995 for test-time tuning.
Batch size was set to 16 for pre-training and to 32 for fine-tuning. Different values depending on the dataset were then used for test-time tuning, see Table \ref{tab:appendix-ablation-hyperparams}. 
Pre-training ran for up to 200 epochs, with early stopping based on validation token accuracy, typically halting around 60 epochs. 
Fine-tuning followed the same criterion, stopping between 20–30 epochs depending on the dataset. 
For test-time tuning, the number of epochs is determined by the number of K-means clusters selected.\\
Teacher forcing technique is implemented for next token prediction~\cite{teacher-forcing}.\\

\begin{figure}[ht]
    \centering
    \vspace{0.5cm}
    \includegraphics[width=0.8\linewidth]{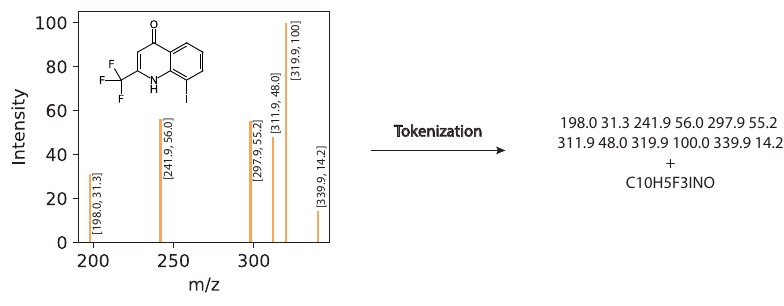}
    \caption{Illustration of the tokenization process for an MS/MS spectrum of the corresponding depicted molecule with formula C$_{10}$H$_5$F$_3$INO. Every peak is stored as a tuple of mass-to-charge ratio and intensity, then converted to a simple string and concatenated with the chemical formula.}
    \label{fig:tokenization}
\end{figure}

\paragraph{Tokenization} To convert MS/MS spectra into a format suitable for the transformer model, we tokenize each spectrum as a sequence of peak pairs $[m/z,I]$, see Figure \ref{fig:tokenization}. 
The peaks are then concatenated into a string representation. This tokenized spectrum is combined with the chemical formula of the respective molecule to form the complete input. By treating both the spectrum and formula as text, the model can leverage language modeling techniques for end-to-end SMILES generation. 
% % Preprocessor details:
% A preprocessor is built for every modality. Different regex expressions are used for Chemical formula and SMILES, as showed in Table \ref{tab:regex}, while the tokenization process for the MS/MS spectra is described in the main text (see Section \ref{sec:methods-data}).

% \begin{table}[ht]
% \centering
%     \begin{tabular}{l|c}
%          & tokenizer\_regex \\ \hline
%        Chemical formula  &  \texttt{([A-Z]{1}[a-z]?[0-9]*)} \\
%        SMILES & \textcolor{blue}{FIX!!!}
%        % \texttt{(\textbackslash[[\XOR\textbackslash ]]+]|Br?|Cl?|N|O|S|P|F|I|b|c|n|o|s|p|\textbackslash(|\textbackslash)|\textbackslash.|=|#|-|\textbackslash+|\textbackslash\textbackslash\textbackslash\textbackslash|\textbackslash/|:|~|@|\textbackslash?|>|\textbackslash*|\textbackslash\$|\textbackslash\%[0-9]{2}|[0-9])} 
%        \\
%     \end{tabular}
%     \label{tab:regex}
% \end{table}

\paragraph{Fingerprint alignment}\label{sec:appendix-model-fing-alignment}
Table \ref{tab:appendix-ablation-alignment} summarizes the performance of pre-trained and fine-tuned models on simulated dataset and NPLIB1 respectively, with and without the use of the loss term based on the fingerprints, described in Section \ref{sec:methods-models}. Different values of the weight $\lambda$ (see Equation \ref{eq:loss}) are explored.
The inclusion of the alignment term improves performances, in particular, when $\lambda$ = 0.1 during pre-training, and even more when $\lambda$ is set to 1 when fine-tuning. 

\begin{equation}
    \mathcal{L} = \mathcal{L}_\text{CE, SMILES} + \lambda \ \mathcal{L}_\text{BCE,Fingerprints}
\label{eq:loss}
\end{equation}

\begin{table}[ht]
    \vspace{0.5cm}
    \centering
    \resizebox{0.8\textwidth}{!}{%
    \begin{tabular}{lcccc}
    \toprule
    & $\lambda$ & Top-1 Accuracy $\uparrow$ & Top-5 Accuracy $\uparrow$ & Top-10 Accuracy $\uparrow$ \\
    \toprule 
    \multirow[c]{3}{*}{\makecell{Pre-training simulations dataset}} 
    & 0.0 & 34.05\% & 57.73\% & 62.69\%\\
    &0.1 & \textbf{36.02\%} & \textbf{59.45\%} & \textbf{64.36\%}\\
    &1 & 34.46\% & 57.40\% & 61.98\%\\
    \midrule
    \multirow[c]{4}{*}{\makecell{Fine-tuning \textsc{NPLIB1-Full}}} 
    & 0.0 & 11.62\% & 25.35\% & 29.88\% \\
    & 0.1 & 11.64\% & 24.99\% & 29.44\%\\
    & 1 & \textbf{12.42\%} & \textbf{25.91\%} & 30.83\% \\
    & 10 & 11.74\% & 25.84\% & \textbf{31.30\%}\\
    \bottomrule
    \end{tabular}%
    }
    \caption{Performances of pre-training on simulated dataset (top section) and fine-tuning on NPLIB1 using different values of $\lambda$, weighting the loss term based on fingerprints prediction.}
    \label{tab:appendix-ablation-alignment}
\end{table}

\section{Additional results}

\subsection{Analysis performances based on specific spectral features}
To better understand the factors affecting the performance of the test-time tuned model on MassSpecGym, we analyzed accuracy and structural validity across subsets grouped by key spectral features: adduct type, instrument, parent mass, and collision energy. The results of this analysis are shown in Table \ref{tab:results-features-msg}.\\
First of all, we divided the dataset depending on the adduct. The majority of spectra (80.12\%) were protonated species (H$^+$), for which the model achieved a Top-1 accuracy of 3.92\%. 
In contrast, spectra with sodium adducts (Na$^+$) represented only 19.88\% of the data and exhibited much lower accuracy (0.09\% Top-1), suggesting that the model struggles with non-protonated species, likely due to their lower representation and more complex fragmentation patterns. The fact that the pre-training simulated data contained only spectra with H$^+$ adduct, definitely contributed as well. Despite this, the model manages to correctly guess a few of them.\\
Secondly, we analyzed the impact of the instrument used to measure the spectra. The ones acquired on Orbitrap instruments dominated the dataset (83.86\%), yielding Top-1 accuracy of 2.86\%. 
QTOF spectra, present with a fraction of 13.81\%, showed higher Top-1 accuracy (5.24\%), indicating that instrument-specific fragmentation characteristics may influence reconstruction quality. 
Information about the instrument is missing for a small portion of the data (2.33\%). The model performed poorly on these across all metrics.\\
In Table \ref{tab:results-features-msg} is also possible to see that performance varied substantially with precursor mass. Compounds in the 400–600 Da range achieved the highest Top-1 accuracy (4.75\%), while smaller molecules (mass within 200-400 Da) achieved slightly lower performances (3.32\% Top-1), but still higher than the accuracy on the whole dataset (3.16\% Top-1).
On the other hand, large molecules (m > 600 Da) were almost never reconstructed correctly, with even 0\% Top-1, 0.34\% valid SMILES for molecules with mass larger than 800 Da. This trend suggests that the model is optimized for mid-range masses, while extreme sizes pose challenges, probably due to increased structural complexity and lower availability.
In the end, we also addressed the impact of the collision energy used during the measurement, which shows to strongly impact accuracy. Intermediate energies ($\sim$ 45–105 $eV$) yielded the best results, whereas very low (<30 $eV$) or very high (>105 $eV$) energies led to lower performance. This indicates that fragmentation richness at moderate energies provides the most informative spectra for structure prediction. The lowest performance is although achieved for the spectra for which we have no information about the collision energy, which constitute a massive part (42.12\%) of the dataset, with a Top-1 accuracy of only 1.08\%.\\ 
Overall, these results highlight that data distribution and fragmentation conditions critically shape model performance, with clear biases toward protonated species, mid-range masses, and spectra acquired under intermediate collision energies.

\begin{table}[h!]
    \centering
    \resizebox{\textwidth}{!}{
    \begin{tabular}{lcccccc}
    \toprule
    % & \multirow[c]{2}{*}{} & \multicolumn{3}{c}{ACCURACY $\uparrow$} & \multirow[c]{2}{*}{VALID $\uparrow$} \\
    % \cmidrule(l{0.15cm}r{0.15cm}){3-5}
    \multicolumn{6}{c}{\textsc{MassSpecGym}} \\
    \toprule 
    & & Portion &\quad Top-1 Accuracy $\uparrow$ \quad &\quad Top-5 Accuracy $\uparrow$ \quad &\quad Top-10 Accuracy $\uparrow$ \quad & Valid SMILES $\uparrow$\\
    \midrule
    \multirow[c]{2}{*}{Adduct} & H$^+$ & 80.12\% & 3.92\% & 6.69\% & 7.52\% & 72.57\% \\
    &Na$^+$ & 19.88\% & 0.09\% & 0.14\% & 0.14\% & 30.76\% \\
    \midrule
    \multirow[c]{3}{*}{Instrument}
    & Orbitrap & 83.86\%& 2.86\% & 4.9\% & 5.52\% & 65.06\% \\
    & QTOF & 13.81\% & 5.24\% & 9.08\% & 10.11\% & 63.77\% \\ 
    & Unknown & 2.33\% & 1.47\% & 1.47\% & 1.47\% & 40.05\% \\ 
    \midrule
    \multirow[c]{4}{*}{Parent mass}
    & $200 \ Da< m < 400 \ Da$ & 21.96\% & 3.32\% & 6.2\% & 7.42\% & 99.74\% \\ 
    & $400 \ Da< m < 600 \ Da$ & 43.06\% & 4.75\% & 8.29\% & 9.18\% & 91.08\% \\ 
    & $600 \ Da< m < 800 \ Da$ & 27.83\% & 1.39\% & 1.66\% & 1.72\% & 11.32\% \\ 
    & $m > 800 \ Da$ & 7.15\% & 0.0\% & 0.0\% & 0.0\% & 0.34\% \\ 
    \midrule
    \multirow[c]{9}{*}{Collision energy}
    & $E<15\ eV$ & 2.75\% & 2.90\% & 5.19\% & 5.39\% & 68.26\% \\ 
    & $15\ eV< E < 30\ eV$  & 17.63\% & 3.65\% & 6.65\% & 7.78\% & 88.80\% \\ 
    & $30\ eV< E < 45\ eV$ & 11.10\% & 5.34\% & 9.96\% & 10.93\% & 84.47\% \\ 
    & $45\ eV< E < 60\ eV$ & 6.90\% & 6.77\% & 11.72\% & 13.28\% & 84.62\% \\ 
    & $60\ eV< E < 75\ eV$  & 14.68\% & 4.23\% & 7.53\% & 8.47\% & 90.94\% \\ 
    & $75\ eV< E < 90\ eV$& 1.59\% & 7.89\% & 15.41\% & 16.49\% & 81.40\% \\ 
    & $90\ eV< E < 105\ eV$& 1.30\% & 6.58\% & 8.77\% & 10.96\% & 82.89\% \\ 
    & $E > 105\ eV$ & 1.95\% & 4.39\% & 4.39\% & 4.39\% & 65.29\% \\ 
    & Unknown & 42.12\% & 1.08\% & 1.45\% & 1.60\% & 34.60\% \\ 
    \bottomrule
    \end{tabular}
    }
    \vspace{-0.2cm}
    \caption{Performances of test-time tuned model on MassSpecGym \cite{bushuiev_massspecgym_2024} subsets obtained grouping the data depending on different features of the spectra. In particular, the performances have been analyzed depending on adduct, collision energy and instrument used to obtain each spectrum, and parent mass of the compound.}
    \label{tab:results-features-msg}
\end{table}

\subsection{Data points selection during test-time tuning}
\begin{figure*}[ht]
\centering
\begin{center}
\includegraphics[width=0.3
\textwidth]{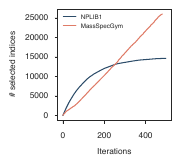}
\vspace{-0.5cm}
\caption{Length of the set of selected indices over the test-time tuning iterations, for NPLIB1-Full and MassSpecGym. In the first case, almost all the points in the candidate pool are selected, showing their relevance for the test points. In the case of MassSpecGym, instead, the method keeps selecting new points at every iteration, because of the larger size of the test set, but also implies diversity among the test points and between the test and train points.}
\label{fig:ttt-set-sel-indices}
\end{center}
\end{figure*}

Figure \ref{fig:ttt-set-sel-indices} illustrates the evolution of the number of selected training points across test-time tuning iterations for NPLIB1 and MassSpecGym. 
In the case of NPLIB1, the curve rapidly approaches the size of the candidate pool ($\sim$ 15 000), indicating that most points are relevant for adaptation in an in-distribution setting. 
Conversely, for MassSpecGym, the selection grows more gradually and continues to increase throughout the iterations, reflecting the diversity and domain shift between training and test sets. This behavior highlights the adaptive nature of the selection process and its dependence on dataset characteristics.

\vspace{-0.5cm}
\subsection{NPLIB1 filtered by H$^+$ adduct}
We report in Table \ref{tab:results-nplib1h} the results of our models on the dataset NPLIB1-H$^+$, obtained as described in Appendix \ref{sec:appendix-data}.

\begin{table}[ht]
    \centering
    \resizebox{0.85\textwidth}{!}{
    \begin{tabular}{lcccc}
    \toprule
    % & \multirow[c]{2}{*}{} & \multicolumn{3}{c}{ACCURACY $\uparrow$} & \multirow[c]{2}{*}{VALID $\uparrow$} \\
    % \cmidrule(l{0.15cm}r{0.15cm}){3-5}
    \multicolumn{5}{c}{\textsc{NPLIB1-H$^+$}} \\
    \toprule 
    &\quad Top-1 Accuracy $\uparrow$ \quad &\quad Top-5 Accuracy $\uparrow$ \quad &\quad Top-10 Accuracy $\uparrow$ \quad & Valid SMILES $\uparrow$\\
    \midrule
    % Fine-tuning from scratch &  0.62\% & \% & 2.09\% & 11.86\% \\ 
    % Zero-shot PT mdoel & 3.83\% & \% & 9.17\% & 74.32\% \\ 
    {Fine-tuning} & {14.11\%} & {27.56\%} & {32.89\%} & {89.24\%} \\
    Test-time tuning& {14.61\%} & {27.19\%} & {32.55\%} & {89.96\%} \\
    % Test-time tuning FT model& {\%} & {\%} & {\%} & {\%} \\
    % {Extended Fine-tuning PT model} & {\%} & {\%} & {\%} & {\%} \\
    % \textbf{Extended Test-time tuning PT model}&  \textbf{\%} & {\%} & \textbf{\%} & {\%} \\ 
    \bottomrule
    \end{tabular}
    }
    \caption{Performances of the fine-tuned and test-time tuned models (starting from the pre-trained model on the simulated dataset) on the experimental dataset NPLIB1-H$^+$, obtained by filtering the original NPLIB1-Full dataset depending on the adduct used to obtain each spectrum, as described in Appendix \ref{sec:appendix-data}.}
    \label{tab:results-nplib1h}
\end{table}

\section{Ablations}
\begin{figure}[ht]
    \centering
    \vspace{-0.2cm}
    \includegraphics[width=0.7\linewidth]{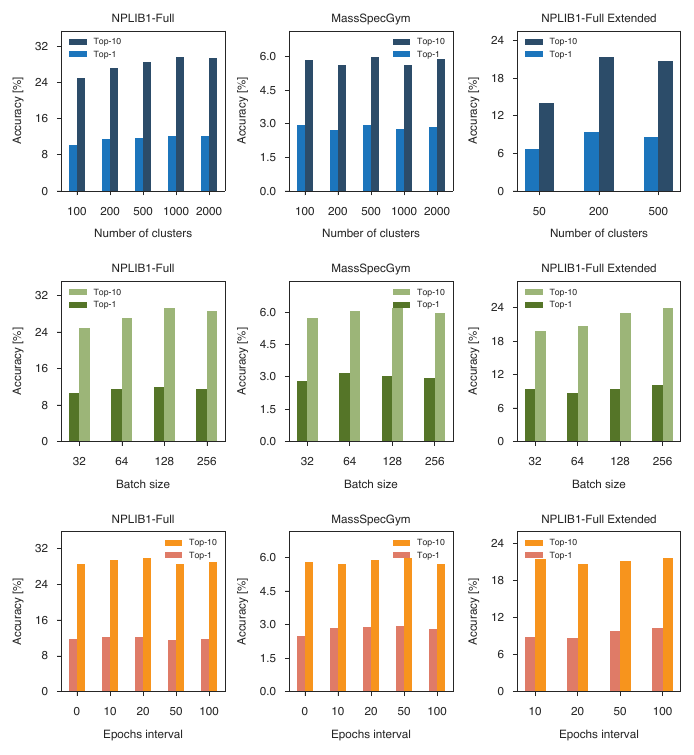}
    \caption{Performances of the test-time tuned models on NPLIB1-Full, MassSpecGym and extended NPLIB1-Full dataset, by varying number of clusters (iterations), batch size (number of training points selected per test point) and number of epochs after which the embeddings are updated.}
    \label{fig:ablations}
\end{figure}

To assess the impact of key hyperparameters on model performance, we conducted ablation studies varying batch size, number of K-means clusters, and number of epochs after which embeddings (and fingerprints) are updated. These experiments were performed on the full versions of the datasets, and the corresponding results are presented in Figure \ref{fig:ablations}. 
Based on these findings, the hyperparameter values selected for batch size, number of K-means clusters, and update frequency in the test-time tuning runs described in the main text are summarized in Table \ref{tab:appendix-ablation-hyperparams}. The best-performing parameters identified through these ablations were subsequently applied to the derived dataset variants (NPLIB1-H$^+$, NPLIB1-DiffMS, and MassSpecGym-DiffMS).

\begin{table}[ht]
    \vspace{0.2cm}
    \centering
    \resizebox{0.55\textwidth}{!}{
    \begin{tabular}{ccccc}
    \toprule
    & {NPLIB1-Full} & {MassSpecGym} & Extended NPLIB1-Full  \\
    \toprule 
    batch size & 128 & 64 & 256 \\
    \# clusters & 1000 & 500 & 1000\\
    \# epochs & 20 & 50 & 100 \\
    \bottomrule
    \end{tabular}
    }
    \caption{Final hyperparameters used for test-time tuning experiments reported in the main text, depending on the dataset. Values were chosen according to the results in Figure \ref{fig:ablations}.}
    \label{tab:appendix-ablation-hyperparams}
\end{table}

